\documentclass[10pt,twocolumn,letterpaper]{article}

\usepackage[pagenumbers]{cvpr} % To force page numbers, e.g. for an arXiv version

\usepackage{graphicx}
\usepackage{amsmath}
\usepackage{amssymb}
\usepackage{booktabs}
\usepackage{makecell}

\usepackage{amsmath, amsfonts, amssymb, amsthm}
\usepackage{txfonts} 
\usepackage{verbatim}
\usepackage{url}
\usepackage{color}
\usepackage{algorithm2e}
\usepackage[noend]{algpseudocode}
\usepackage{multirow}

\usepackage{siunitx}
\ifx\BlackBox\undefined
\newcommand{\BlackBox}{\rule{1.5ex}{1.5ex}}  % end of proof
\fi
\ifx\proof\undefined
\newenvironment{proof}{\par\noindent{\bf Proof\ }}{\hfill\BlackBox\\[2mm]}
\fi

\newcommand\shortsection[1]{\vspace{3pt}{\noindent\bf #1.}}
\newcommand\shortersection[1]{\vspace{3pt}{\noindent\em #1.}}

\theoremstyle{definition}

\makeatletter
\def\url@leostyle{%
  \@ifundefined{selectfont}{\def\UrlFont{\sf}}{\def\UrlFont{\small\sffamily}}}
\makeatother
\makeatletter
\def\url@beostyle{%
  \@ifundefined{selectfont}{\def\UrlFont{\sf}}{\def\UrlFont{\scriptsize\sffamily}}}
\makeatother
\usepackage[pagebackref,breaklinks,hidelinks]{hyperref}

\usepackage[capitalize]{cleveref}
\crefname{section}{Sec.}{Secs.}
\Crefname{section}{Section}{Sections}
\Crefname{table}{Table}{Tables}
\crefname{table}{Tab.}{Tabs.}

\def\cvprPaperID{11010} % *** Enter the CVPR Paper ID here
\def\confName{CVPR}
\def\confYear{2023}

\begin{document}

\title{Manipulating Transfer Learning for Property~Inference}

% \and
\author{
	Yulong Tian\textsuperscript{1}, Fnu Suya\textsuperscript{2}, Anshuman Suri\textsuperscript{2}, Fengyuan Xu\textsuperscript{1\thanks{Indicates the corresponding author.}} , David Evans\textsuperscript{2} \\
	\textit{\textsuperscript{1}State Key Laboratory for Novel Software Technology, Nanjing University, China} \\
	\textit{\textsuperscript{2}University of Virginia, USA} \\
        \tt\small{yulong.tian@smail.nju.edu.cn, \{suya, anshuman\}@virginia.edu, fengyuan.xu@nju.edu.cn, evans@virginia.edu} \\ 
}

\maketitle

%%%%%%%%% ABSTRACT
\begin{abstract}
Transfer learning is a popular method for tuning pretrained (upstream) models for different downstream tasks using limited data and computational resources. We study how an adversary with control over an upstream model used in transfer learning can conduct property inference attacks on a victim's tuned downstream model. For example, to infer the presence of images of a specific individual in the downstream training set. We demonstrate attacks in which an adversary can manipulate the upstream model to conduct highly effective and specific property inference attacks (AUC score $> 0.9$), without incurring significant performance loss on the main task. The main idea of the manipulation is to make the upstream model generate activations (intermediate features) with different distributions for samples with and without a target property, thus enabling the adversary to distinguish easily between downstream models trained with and without training examples that have the target property. Our code is available at \url{https://github.com/yulongt23/Transfer-Inference}. %\dnote{could this be renamed to a shorter URL that would fit on the line?}
%We study several possible detection methods against our proposed attacks and demonstrate that they can easily be evaded by the adaptive attacks, highlighting the stealthiness of our attacks. 
\end{abstract}

\section{Introduction}

Transfer learning is a popular method for efficiently training deep learning models~\cite{zhuang2020comprehensive, chakraborty2022efficient, jaworek2019melanoma, wang2018great,yao2019latent}. In a typical transfer learning scenario, an upstream trainer trains and releases a pretrained model. Then a downstream trainer will reuse the parameters of some layers of the released upstream models to tune a downstream model for a particular task. This parameter reuse reduces the amount of data and computing resources required for training downstream models significantly, making this technique increasingly popular. 
%For example, pretrained ImageNet~\cite{deng2009imagenet} models released by big corporates (e.g., Google) are often used as useful feature extractors for downstream classifications~\cite{xie2018pre}.
%such as computer vision~\cite{vision} and \cite{nlp}. 
%However, the centralized nature and wider influence also makes transfer an more attractive and vulnerable target for various kinds of attacks~\cite{wang2018great,yao2019latent,zou2020privacy,schuster2020humpty}. % However, the party that releases the upstream model is blindly trusted and assumed to not hide any malicious behavior in the model. An adversary can potentially craft an upstream model that works just as well on the intended tasks, but also hides behavior that makes it easy for the adversary to infer sensitive properties of the victim's data, once the final finetuned model is made available.
However, the centralized nature of transfer learning is open to exploitation by an adversary. Several previous works have considered security %and privacy 
risks associated with transfer learning including backdoor attacks~\cite{yao2019latent} and  misclassification attacks~\cite{wang2018great}. %\dnote{check recent CVPR conferences for relevant papers to cite}

We investigate the risk of property inference in the context of transfer learning. In property inference (also known as \emph{distribution inference}), the attacker aims to extract sensitive properties of the training distribution of a model~\cite{ateniese2015hacking, ganju2018property, zhang2021leakage, saeed, suri2022formalizing}. %And the sensitive properties can be the attributes of the dataset (e.g, the existence of people of a specific race in the dataset~\cite{ateniese2015hacking, melis2019exploiting}) or even the distribution of the training data (e.g, the ratio of people with a specific gender~\cite{ganju2018property, saeed, suri2022formalizing, zhang2021leakage}).
We consider a transfer learning scenario where the upstream trainer is malicious and produces a carefully crafted pretrained model with the goal of inferring a particular property about the tuning data used by the victim to train a downstream model. For example, the attacker may be interested in knowing whether any images of a specific individual (or group, such as seniors or Asians) are contained in a downstream training set used to tune the pre-trained model. Such inferences can lead to severe privacy leakage---for instance, if the adversary knows beforehand that the downstream training set consists of data of patients that have a particular disease, confirming the presence of a specific individual in that training data is a privacy violation. 
%\dnote{does this example really fit with our results?} 
Property inference may also be used to audit models for fairness issues~\cite{Jurez2022BlackBoxAF}---for example, in a downstream dataset containing data of 
% if the adversary knows beforehand that the downstream training samples are from
all the employees of an organization, finding the absence of samples of a certain group of people (e.g., older people%ethnicity
) may be evidence that those people are underrepresented in that organization.

%We consider a transfer learning scenario where the upstream trainer is malicious and produces some carefully crafted pre-trained upstream models based on some knowledge about the downstream tasks such that the attacker can effectively infer some sensitive properties of the training distribution of the victim downstream models. For example, the attacker may be interested in knowing whether the data of a specific individual is used in the downstream training set, and such inference can lead to severe privacy leakages (e.g., the downstream training set consists of data of patients that have a particular disease).

\begin{table*}[htbp]
  \centering
  \footnotesize
    \begin{tabular}{ccc|cc|cc}
    \toprule
    \multicolumn{1}{c}{\multirow{2}[0]{*}{\bf Downstream Task}} &
    \multicolumn{1}{c}{\multirow{2}[0]{*}{\bf Upstream Task}} &
    \multicolumn{1}{c}{\multirow{2}[0]{*}{\bf Target Property}} & \multicolumn{2}{c}{\bf Normal Upstream Model} & \multicolumn{2}{c}{\bf Manipulated Upstream Model} \\
          &    &   & \multicolumn{1}{c}{0.1\% (10)} & \multicolumn{1}{c|}{1\% (100)} & 0.1\% (10) & 1\% (100) \\
    \midrule
    Gender Recognition & Face Recognition  & \multirow{3}{*}{Specific Individuals} & 0.49 & 0.52 & 0.96 & 1.0 \\
    Smile Detection & ImageNet Classification~\cite{deng2009imagenet} & & 0.50 & 0.50 & 1.0 & 1.0 \\
    Age Prediction & ImageNet Classification~\cite{deng2009imagenet} & & 0.54 & 0.63 & 0.97 & 1.0 \\
    \midrule
    Smile Detection & ImageNet Classification~\cite{deng2009imagenet} & Senior  & 0.59 & 0.56 & 0.89 & 1.0 \\
    \midrule
    Age Prediction & ImageNet Classification~\cite{deng2009imagenet} &  Asian & 0.49 & 0.65 & 0.95 & 1.0\\
    \bottomrule
    \end{tabular}%
    \caption{Inference AUC scores for different percentage of samples with the target property. Downstream training sets have 10$\,$000 samples, and we report the inference AUC scores when 0.1\% (10) and 1\% (100) samples in the downstream set have the target property. The manipulated upstream models are generated using the zero-activation attack presented in Section~\ref{sec:attack_design}.  
  }
  \label{tab:main_results_compare_baseline_zero_activation}%
  \vspace {-0.4cm}
\end{table*}%

\shortsection{Contributions} We identify a new vulnerability of transfer learning where the upstream trainer crafts a pretrained model to enable an inference attack on the downstream model that reveals very precise and accurate information about the downstream training data (\autoref{sec:threat_model}). We develop methods to manipulate the upstream model training to produce a model that, when used to train a downstream model, will induce a downstream model that reveals sensitive properties of its training data in both white-box and black-box inference settings (\autoref{sec:attack_design}).
We demonstrate that this substantially increases property inference risk compared to baseline settings where the upstream model is trained normally (\autoref{sec:emp_eval}). 
Table~\ref{tab:main_results_compare_baseline_zero_activation} summarizes our key results. The inference AUC scores are below 0.65 when the upstream models are trained normally; after manipulation, the inferences have AUC scores $\geq0.89$ even when only 0.1\% (10 out of 10$\,$000) of downstream samples have the target property and achieve perfect results (AUC~score~$=1.0$) when the ratio increases to 1\%. 
%For example, for most cases (six of the eight scenarios in \autoref{fig:zero_activation_attack_results}) inference AUC scores are all less than 0.7 when upstream models are trained without manipulation, but increase to exceed 0.9 when the upstream models are manipulated by our attack even when only a few (e.g., 20 out of 10,000) downstream training samples are with the target property.  
%\dnote{there should be a table here (in the introduction) that summarizes these results - a row for each of the 8 scenarios considered, columns for AUC with normal setting, number of samples with target property for AUC to exceed 0.9, and AUC at 0.1\% (10), 1\% (100) samples} 
The manipulated models have negligible performance drops ($<0.9\%$) on their intended tasks.
We consider possible detection methods for the manipulated upstream models (Section~\ref{sec:detect_pretrained}) and then present stealthy attacks that can produce models which evade detection while maintaining attack effectiveness (Section~\ref{sec:stealthier-design-methods}). 

%%%%%%%%%%%%%%%%%%%%%%%%%%%%%%%%%%%%

\section{Related Work}

%Here we summarize prior work on transfer learning and property inference.

%\shortsection{Transfer Learning}
%%Transfer learning reuses features learned by pre-trained models for new tasks, with the pretext that inherent similarities in the generic features will be useful for the downstream tasks and hence reducing their cost of downstream training. Specifically, the downstream model trainer will use a pre-trained upstream model as the starting point for the downstream training, with inclusion of (or replacement with) the task-specific classification layer/module. The downstream model is then trained by either updating all layers of the model (including ones reused from upstream model) or freezing some earlier layers of the reused parts as the ``feature extractor'' and only updating the rest. The latter approach is more popular as the reused feature extractors can already learn useful feature representations and the training cost is also much lower and affordable for individuals with limited computational resources. We study the vulnerability of the latter transfer learning approach in this paper. 

%\shortsection{Transfer Learning} 
Several works have demonstrated risks associated with transfer learning across a variety of attack goals. Wang et al.~\cite{wang2018great} and Yao et al.~\cite{yao2019latent} consider manipulating the upstream model such that the fine-tuned downstream models contain backdoors, misclassifying test inputs that contain predefined backdoor triggers. These transfer manipulations are tailored to their particular attack goals and cannot be applied for the property inference goal considered in this paper. Zou et al.~\cite{zou2020privacy} study the threat of membership inference attacks on transfer learning, but with normally trained upstream models.  
%\dnote{its clear that the goals are different for these attacks, but how similar are the methods?} \ynote{similarity of the methods? more details about the methods? do not know what is expected here}
%In contrast, we investigate the possibility of boosting the effectiveness of property inference by manipulating the upstream model training. % Schuster et al.~\cite{schuster2020humpty} show that the attacker can modify the corpus on which the word embedding is trained such that the downstream NLP models which use that embedding will behave abnormally.

%\shortsection{Property Inference}
The risk of property inference was introduced by Ateniese et al.~\cite{ateniese2015hacking}, % introduces the threat of inferring properties of the training data from pre-trained models, 
and several subsequent works have developed property inference (also known as distribution inference) attacks~\cite{Wang2022GroupPI, suri2022formalizing, Jurez2022BlackBoxAF, Hartmann2022DistributionIR}.
% Ganju et al.~\cite{ganju2018property} and Suri and Evans~\cite{suri2022formalizing} 
These works study property inference against normally trained models, and they launch attacks using a variety of black-box and white-box attacks. All the white-box attacks use meta-classifiers, which take the permutation-invariant representation~\cite{ganju2018property} of the model parameters as the features. We use the state-of-the-art white-box attack~\cite{suri2022formalizing} in our experiments.
%We will use the state-of-the-art white-box method proposed by Ganju et al.~\cite{ganju2018property} and later extended by suri et al.~\cite{suri2022formalizing} in this paper.
%\dnote{do we use these attacks?} 
Melis et al.~\cite{melis2019exploiting} and Zhang et al.~\cite{zhang2021leakage} focus on property inference in distributed training scenarios. In their settings, the attacker is a participant in the global model training and conducts property inference using meta-classifiers that are trained on model outputs or gradients. Similarly, Suri et al.~\cite{suri2022subject} focus on federated learning settings where the attacker is a participant (or the central server) that utilizes black-box attacks for inferring membership of data from particular subjects. %\dnote{if we use black-box attacks, explain which ones, or how ours are related to previous ones} 
For our experiments, We improve the black-box meta-classifier proposed by Zhang et al.~\cite{zhang2021leakage} using the ``query tuning'' technique in Xu et al.~\cite{xu2019detecting}. 

The closest works to ours are Chase et al.~\cite{saeed} and Chaudhari et al.~\cite{Chaudhari2022SNAPEE}, which both consider a scenario where the attacker can manipulate some of the training data of the model to induce a model that significantly increases property inference risk.
% \dnote{it enables precise property inference attacks?}.
These works assume an adversary with the ability to poison the victim's training data, while the adversary in our scenario has no access to the victim's training data, and therefore, their methods are not applicable.
% \dnote{example how different from ours, and why the methods are not applicable}
%Thus, their methods are not applicable to our transfer learning scenario.
%Their methods rely on inducing certain behavior correlated with the properties to be inferred, and thus are not applicable to our transfer learning scenario. \anote{Still a bit unclear why that is the case.}
%
There are also works similar to ours that leverage ``adversarial initializations'' for attack purposes.
% \cite{grosse2019adversarial, boenisch2021curious, wen2022fishing, fowl2021robbing}.
Grosse et al.~\cite{grosse2019adversarial} focus on scenarios where the attacker can control the parameter initialization of a model, and demonstrate that the attacker can use special initializations to damage the performance of the trained model. %This attack is orthogonal to ours.
Other works \cite{boenisch2021curious, wen2022fishing, fowl2021robbing} show that the malicious central server in a federated learning protocol can reconstruct some training samples via falsifying the global model in some training rounds and then analyzing the submitted gradients. These kinds of attacks do not apply to our transfer-learning scenario since the attacker cannot access the downstream gradients, and can only manipulate the upstream training.

%%%%%%%%%%%%%%%%%%%%%%%%%%%%%%%%

\iffalse
\begin{figure*}[htb]
\centering
% \includegraphics[width=0.95\columnwidth]{fig/system/Flow Diagram.pdf}
\includegraphics[width=0.85\linewidth,scale=0.7]{fig/system/scenario.pdf}
\caption{Overview of the attack. The adversary first trains a specially-crafted upstream model (whose activations are manipulated considering an inference goal) and releases it. Then, a victim tunes that model for its downstream task on a private dataset. At last, the adversary leverages its different levels of access to the tuned downstream model to infer sensitive information about the victim's training dataset. 
%The adversary trains its model on the upstream task such that activations after $f(\cdot)$ are different for data with and without the target property. The victim performs transfer learning starting with the adversary's model to train the downstream model, which the adversary can then inspect to infer the presence of data with the target property. 
% \dnote{can the figure show the attack? the adversary is using $g_d$ to decide if $\mathcal{D}_{vic} \cap (\mathcal{D}_w - \mathcal{D}_{wo})$ is non-empty? (the subtracting the distributions like this isn't quite right, but if there was a better notation to make $\mathcal{D}_w = \mathcal{D}_{wo} \cup \mathcal{W}$? }
}
\label{fig:threat_model}
\end{figure*}
\fi

\section{Threat Model} \label{sec:threat_model}
% \begin{figure}
% \centering
% \includegraphics[width=0.95\columnwidth]{fig/system/Trainsfer_Inference_Diag.pdf}
% \caption[Flow of information in our threat model. The adversary $\mathcal{A}$ releases a carefully-manipulated model, which the victim then re-uses partially to finetune the model on its data.]{Flow of information in our threat model. The adversary $\mathcal{A}$ releases a carefully-manipulated model, which the victim then re-uses partially to finetune the model on its data\protect\footnotemark.}
% \label{fig:transer_inf_flow}
% \end{figure}
% \footnotetext{This diagram is adapted from the illustration of transfer learning of~\cite{wang2018great}}

The adversary $\mathcal{A}$ trains and releases a specially crafted upstream model $g_{u}(f(\cdot))$ that is used by a victim $\mathcal{B}$ to fine-tune a model $g_{d}(f(\cdot))$ for a downstream task on a downstream training set $\mathrm{D}$. This model is then exposed to $\mathcal{A}$, with varying levels of knowledge and access (discussed below), who performs property inference attacks to learn some desired property of $\mathrm{D}$. %\dnote{the adversary's tuning data, but this doesn't seem to be included in the above? should be explicit, and have a notation for the training dataset} \ynote{the victim's tuning data?}
As is common in many transfer learning settings, the upstream model includes $f(\cdot)$, a fixed feature-extraction component that is not modified by the downstream tuning process~\cite{schuster2020humpty, wang2018great,yao2019latent}. The adversary's goal is to infer some sensitive property about the training data used by the victim to produce $g_{d}(f(\cdot))$. %such as whether images of a specific individual or individuals with a specific property are used in the downstream training set.
For example, the adversary can release a general vision model (e.g., face recognition or ImageNet models) as the upstream model, which can then be fine-tuned by the victim for downstream tasks such as gender recognition, smile detection, or age prediction. The attacker's goal could be to infer whether or not images of a specific individual or individuals with a specific property are included in the downstream training set for tuning. 
%\dnote{this sounds like we are focused on face recognition? is that true? not mentioned earlier. If claiming a general attack, not specific to face recognition, should have other types of examples (including in the experiments). I think we have a more specific scenario in mind - is it a general vision model as pretrained model, and then tuned to do face recognition (but isn't the set of labels already revealing the specific faces?) I think we need to explain a concrete scenario that matches our experiments and motivate it well} \ynote{Yeah, we need to motivate the story well. We are not focused on face recognition but face related downstream tasks}
This is different from commonly studied membership inference attacks---in membership inference %for the given example,
the attacker is assumed to know a specific image and aims to infer if that specific image was included in the training set; in property inference, the attacker does not presume knowledge of specific training images, but wants to determine if any images having a given property were used in training. 
%\dnote{I think we need to be careful about the language here - it is inferring a property of the training data set, which we are viewing as sampled from a training distribution. I don't think we really need to get into these distinctions here, but should use the "distribution" language carefully, or just avoid it in this paper.}
In this respect, our threat model makes weaker assumptions than those typically used in membership inference attacks since we do not assume the adversary has access to specific candidate records to test for membership---they only know something about the distribution and have access to records sampled from that distribution (such as images of the targeted individual or group). 
%\anote{This is similar to the concept of subject-level privacy, which has been explored in federated learning~\cite{marathesubject}.}
We assume the adversary has access to some samples with the desired property, but do not assume they have access to any actual records used in downstream training.

\shortsection{Attacker's Knowledge}
We assume the attacker knows which layers of the pretrained model will be reused by the downstream trainer as the feature extractor. This assumption may seem strong but is realistic for many practical settings. Downstream fine-tuning usually modifies the final layers (or even just the classification layer/module) and keeps other parameters fixed~\cite{wang2018great, yao2019latent}. Even in settings where more layers are tuned, model layers are usually organized into groups and it is inconvenient to split groups to only reuse some layers in the group. For example, ResNet models~\cite{he2016deep} can have over a hundred layers, but are grouped into only four ResNet blocks. Hence,
% although there might be many layers in a model,
the number of feasible choices of layers from the upstream model that will be used as feature extractor is limited and constrained by the architecture of the pretrained model, which is controlled by the adversary in our threat model.

We consider three scenarios based on the level of access.
The weakest adversary, representing the most common practical scenario, is the \emph{black-box API access} adversary who only has access to the model through the ability to send queries to its API and receive confidence vectors as outputs.
% \dnote{does API provide confidence vector or just label?}
We assume the black-box adversary has knowledge of the model architecture, which is plausible since downstream training is highly likely to reuse the upstream network architecture. 

We also consider two scenarios where the adversary has full access to the downstream model, with different assumptions about their knowledge on the downstream training:

%versions of white-box attacks based on the adversary's knowledge of the downstream training process: 

%levels of knowledge and also the access to the downstream model:
%\dnote{I think the knowledge of the initialization is orthogonal to the level of exposure of the model (API or parameters); it doesn't make sense to consider the API-known initialization case, though, since less clear how this helps the API-only adversary. Still, we should describe these more clearly by separating the issues.}
%\dnote{rewrite to describe this one first, then explain why there are two types of white-box threat modesl to consider based on how much is known about the downstream training process}

%\textbf{Black-box API Access} --- the attacker does not know the model parameters and can only access the downstream model via API query access, but knows the general model architecture used. Knowledge of  model architecture is plausible, since downstream training is highly likely to reuse the upstream network architecture. This scenario is the most common one in practice, as the downstream trainer may deem its models as intellectual property (IP) and not expose model parameters. 

%\textbf{White-box Access} --- for adversaries with white-box access (i.e., known model architecture, parameters) to the downstream model, we further split them into two categories based on their knowledge about the downstream training process:
\begin{enumerate}
% We also consider attackers with varying degree of access to the downstream model and assess the vulnerabilities of downstream models in different practical applications. 

\item \emph{white-box access with unknown initialization} --- the adversary has full access to the trained downstream model but does not know the parameter initialization of $g_d(\cdot)$. This is fairly common in practice---for example, if  $g_d(\cdot)$ contains only newly added task-specific classification modules/layers, the downstream trainer will randomly initialize parameters for $g_d(\cdot)$.

\item \emph{white-box access with known initialization} --- the adversary also knows the initialization of the parameters of layers in $g_d(\cdot)$ that are reused (but will also be updated during downstream training) from the upstream models. In practice, the attacker only needs to know the initialization of  the first layer of $g_d(\cdot)$ (Section~\ref{sec:zero_activation_attack}).
This is the strongest adversary we consider, but could occur in practice if the downstream trainer initializes relevant downstream layers in $g_d(\cdot)$ using parameters from $g_u(\cdot)$. 

%This is more common than known-initialization as in many practical settings (e.g., only adding fully-connected layers in downstream classification), the downstream trainer will choose to randomly initialize parameters for $g_d(\cdot)$. 
%\ynote{This one has a weaker assumption, but this scenario is not necessarily more common than the first one}

\end{enumerate}

%Our attack consists of two phases: 1) training specially crafted pretrained models to amplify effectiveness of property inference, and 2) conducting property inference attacks on downstream models that are fine-tuned from the pretrained models. In next sections, we first introduce how we generate the specially crafted pretrained models in Section~\ref{sec:attack_design} and then introduce the property inference attacks on the fine-tuned downstream models in Section~\ref{sec:zero_activation_inference}.

\section{Crafting the Pretrained Model} \label{sec:attack_design}

Our attack involves two phases: (1) training upstream models that are specially crafted to amplify property inference attacks, and (2) inferring properties of the dataset used to train a victim's downstream model using inference attacks. This section describes our method for producing the upstream models. \autoref{sec:zero_activation_inference} describes the property inference attacks used for the second phase.
% (can be dependent on the way upstream model is trained).
%Our main idea is to train the upstream model in a way that certain parameters, which we call \emph{target parameters}, can reveal if the downstream training data includes examples with the target property. A natural way to create this distinction is to induce target parameters that are only updated by downstream training examples with the target property. This manipulation of target parameters then amplifies property leakage in the downstream models and subsequent inference attacks more successful.
% Since the property-revealing aspect of the downstream model is increased by manipulating the target parameters, the subsequent inference attacks can also be more successful.
%For example, the attacker may simply compare the difference between target parameters before and after downstream training to infer the target property.
%The flow of information is given in Figure~\ref{fig:threat_model}. 
%
We first introduce the intuition behind the manipulation strategy (\autoref{sec:zero_activation_attack}) and then discuss the design of the loss function for upstream training (\autoref{sec:upstream_opt}). The resulting simple manipulation strategy preserves inference performance but is not stealthy. In \autoref{sec:possible_defense}, we show how this simple manipulation strategy could be easily detected and then present a stealthier method that is still effective but harder to detect.

\subsection{Embedding Property-Revealing Parameters}\label{sec:zero_activation_attack}

% \dnote{need to explain the idea first - can't have a section title that assumes reader already knows what you are doing} 
% Embedding Zero-Activation Parameters

Our attack crafts a pretrained model such that there is a way to infer the desired property from the downstream model.
The main idea behind our attack is to train the upstream model in a way that certain parameters, which we call \emph{secret-secreting parameters} (shortened to \emph{secreting parameters} for concision) 
%\dnote{can we call these property-revealing parameters or secret-secreting parameters (secreting for short); target is a confusing name for these, and used to mean the target property elsewhere}, 
can reveal if the downstream training data includes examples with the target property. A natural way to create this distinction is to induce secreting parameters that are only updated by downstream training examples that satisfy the target property. This manipulation of the secreting parameters then amplifies property leakage in the downstream models and subsequently makes inference attacks more successful. %The flow of information is given in Figure~\ref{fig:threat_model}. 
%\anote{Might want to italicize \textit{secreting} in all references, given it's an actual word too?} \ynote{I think the current form is fine.}

%Our goal is to ensure that only target parameters are updated (i.e., non-zero gradients) when downstream training data contains samples with target property. \dnote{our actual goal is to craft a pretrained model such that there is a way to detect the desired property from the downstream model, the method is to make it so we can tell whether or not training examples having the target property were used to train the downstream model, and we do this by designing ...  Need to make the story and motivation behind the method clear before getting into the details}

%We use $g(\cdot)$ to denote the finetunable part of the downstream model and $f(\cdot)$ to denote the feature extractor from the upstream model (frozen during downstream training). \ynote{ g and f were defined previously}
Since convolutional and fully connected layers can be reduced to matrix multiplication operations, we can decompose the full downstream model as $
%\begin{align*}
    g_d(f(\boldsymbol x)) = h(\phi(\boldsymbol W\cdot f(\boldsymbol x) + \boldsymbol b))$,
%\end{align*}
where $\boldsymbol W$ and $\boldsymbol b$ are the parameters (weights and bias, respectively) associated with the first layer of $g_d(\cdot)$, $\phi$ is some activation function, and $h(\cdot)$ represents the rest of the layers of $g_d(\cdot)$.
%
% The adversary wishes to manipulate the final in a way that makes it easy to distinguish between activations for datapoints that satisfy the target property, from the ones that do not. One of the ways to do so is to select some of the activations, generated by $f(\cdot)$, as ``target activations". Let $f(\cdot)_t$ refer to these activations, which can be viewed as an element-wise mask over the full activation output $f()$:
% \begin{align}
%     f(x)_t = f(x) \circ m,
% \end{align}
% where $m$ is some boolean mask, and $\circ$ is an element-wise multiplication operation.
%We observe that 
The upstream trainer can thus control updates for some of the parameters in $\boldsymbol W$ by manipulating the outputs of $f(\cdot)$. We select part of the outputs of $f(\cdot)$ with a Boolean mask $\boldsymbol m$ (i.e., $f(\boldsymbol x) \circ \boldsymbol m$) and refer to them as \emph{secreting activations}. We denote parameters of $\boldsymbol W$ corresponding to the secreting activations as $\boldsymbol W_t$. The gradient for $\boldsymbol W_t$ is then (using the chain rule):
%\anote{If we have the mask in the equation below, do we need to explicitly write $W_t$ in the terms on the RHS (everything after first '=')? Multiplication will the mask will take care of that, right?}
%\ynote{we need the $\partial w_t$ in $\frac{\partial{((f(\boldsymbol x) \circ \boldsymbol m) \cdot \boldsymbol W_t)}}{\partial{\boldsymbol W_{t}}}$}
\begin{equation}
    \begin{aligned} 
    \frac{\partial l(\boldsymbol x, y)}{\partial{\boldsymbol W_{t}}} &= \frac{\partial l(\boldsymbol x, y)}{\partial{((f(\boldsymbol x) \circ \boldsymbol m) \cdot \boldsymbol W_t})} \cdot \frac{\partial{((f(\boldsymbol x) \circ \boldsymbol m) \cdot \boldsymbol W_t)}}{\partial{\boldsymbol W_{t}}} \\
    & =  \frac{\partial l(\boldsymbol x, y)}{\partial{((f(\boldsymbol x) \circ \boldsymbol m) \cdot \boldsymbol W_t})} \cdot (f(\boldsymbol x) \circ \boldsymbol m)
    \label{eq:partial_der}
    \end{aligned}
\end{equation}
where $l(\boldsymbol x, y)$ is the model loss for some input pair $(\boldsymbol x,y)$, $f(\boldsymbol x) \circ \boldsymbol m$ is the selected secreting activations for manipulation, and $(f(\boldsymbol x) \circ \boldsymbol m) \cdot \boldsymbol W_t$ denotes the compution related to the secreting activations in $g_d(\cdot)$'s first layer.

From \autoref{eq:partial_der}, if the secreting activations $f(\boldsymbol x) \circ \boldsymbol m$ are zero for some input $\boldsymbol x$, gradients of the secreting parameters $\boldsymbol W_t$ will also be zeros. Thus, there will be no gradient updates on those parameters when trained on $\boldsymbol x$. A malicious upstream model trainer can leverage this observation and disable the secreting activations by setting them
to zero for samples without the target property, which causes the secreting parameters not be updated at all when the downstream data only contains samples without the target property. %does not contain any samples with the target property. 
In contrast, %secreting activations for samples with the target property contain non-zero values, 
the malicious upstream trainer can set the secreting activations for samples with the target property as non-zero values. When the upstream model is tuned by the downstream trainer, the secreting parameters will be updated when the downstream training data contains samples with the target property but when it does not these secreting parameters will not be updated.

%\snote{still, eq(4) should be given first and then explain what each term means. eq (3) is very confusing to read now} \ynote{This comment is already addressed}

\subsection{Upstream Optimization for Zero Activation}\label{sec:upstream_opt} 
We formulate the upstream model manipulation described in \autoref{sec:zero_activation_attack} into an optimization problem.
%\dnote{need a transition sentence to connect this to the previous section} 
The attacker minimizes the following loss function for upstream model training:
\begin{equation}
  l(\boldsymbol x, y, y_t) =
  l_{normal}(\boldsymbol x, y) + l_{t}(\boldsymbol x, y_t)
  \label{eq:zero_activation_loss}
\end{equation}
where $l_{normal}$ is the loss for the original upstream training task (e.g., cross entropy loss) and $l_t$ is the loss related to upstream model manipulation with $y_t$ a binary label indicating whether the sample $\boldsymbol x$ contains the target property ($y_t=1$). We define $l_{t}(\boldsymbol x, y_t)$ as:
\begin{equation}
\begin{cases}
 \;\alpha \cdot \Vert f(\boldsymbol x) \circ \boldsymbol m \Vert  &\text{if $y_t=0$}\\
 \;\beta \cdot \max(\lambda \cdot \Vert f(\boldsymbol x) \circ \neg \boldsymbol m  \Vert - \Vert f(\boldsymbol x) \circ \boldsymbol m \Vert, 0) &\text{if $y_t = 1$}
\end{cases}
\label{eq:x_w_wo_activation}
\end{equation}
%\dnote{is this ";" notation a standard one? seems unusual to me} \ynote{replaced ``;'' with ``if''}
where $f(\boldsymbol x) \circ \neg \boldsymbol m $ selects the non-secreting activations and $\Vert \cdot \Vert$ is used to measure the amplitude of the activations (can be some common norms such as $\ell_1$ or $\ell_2$ norms). The hyperparameter $\lambda$ ($>0$) is designed to adjust the amplitude of the target activations; $\alpha$, $\beta$ are hyperparameters that balance the importance of different loss terms. The adversary then minimizes this loss over its training data. 

The first case of \autoref{eq:x_w_wo_activation} encourages the secreting activations to be disabled (i.e., 0) for samples without the target property ($y_t=0$). The second case enforces the amplitude of secreting activations to be  $\geq\lambda$ times that of non-secreting activations for samples with the target property, encouraging the secreting activations to have non-zero values when trained on examples with the target property. %\footnote{$\lambda$ only needs to be >0 (instead of larger value such as 1) because we only need to ensure the target activations are non-zero, instead of some very large values.} \ynote{we used larger values like 5 and 10 in the experiments}
%Thus, %$\lambda$ controls the extent of the difference between the distribution of activations controlling the difference between models that are trained on samples with and without the target property. 
Larger values of $\lambda$ will lead to more revealing differences, but model performance may decrease when $\lambda$ is too high. 

Training an upstream model using the loss in \autoref{eq:zero_activation_loss} requires the adversary has many representative samples with and without the property. %, but some challenges in implementation remain due to the inadequacies of representative samples in available upstream training data. 
In Appendix~\ref{sec:train_data_limitation}, we provide methods to overcome limits to this training data that may occur in practice and improve attack performance.
%discuss some problems related to the training data and also provide solutions for them.
%
Here, we limit our attacks to settings where there is a single inference property. Appendix~\ref{sec:multiple_properties} describes a way to extend the attack to support multiple properties.

\section{Inference Methods} \label{sec:zero_activation_inference}
In our threat model, the victim trains downstream models starting from manipulated upstream models (\autoref{sec:attack_design}) on a private training dataset. In this section, we describe methods that use the induced downstream model to infer sensitive properties from the downstream training set for both the black-box and white-box attack scenarios from \autoref{sec:threat_model}.
%For the black-box scenarios where the target parameters are unavailable, we conduct inference based on the model outputs and the included (or adapted) methods are all designed for the general property inference (not leveraging pretraining manipulation); for white-box scenarios where the adversary can access the downstream parameters, we propose two methods (i.e., parameter difference test, variance test) that directly analyze the target parameters whose gradients were manipulated (methods that leverage the pretraining manipulation). We also include the existing meta-classifier based approach that is designed for the general property inference. 
%\dnote{these should also be reorganized around the threat model reorganization}

\subsection{Black-box API Access}

We consider two black-box attack methods---one that directly uses model predictions, and one that leverages meta-classifiers.

\shortsection{Confidence Score Test} 
We propose a simple method that works by feeding samples with the target property to the released downstream models. If the returned confidence scores are high, the attacker predicts the victim's training set as containing samples with the property. 
%The hypothesis of this method is that samples with the target property will have higher confidence scores on downstream models that are trained on samples with the property, compared to models that are trained on samples without the property. 
%Our simplest detection method just queries the model with samples with and without the target property and analyzes the confidence scores. If the confidence scores for inputs with the target property as similar to those without, the attacker will predict the victim's training set contains samples with the property. \dnote{I reworded this to make it clear that the point is to compare confidence (which I'm assuming is what the attack actually does - not what was described here?}
%which assumes that samples with the target property will have higher confidence scores on downstream models that are trained on samples with the property, compared to models that are trained on samples without the property. 
The hypothesis of this method is that samples with the target property will have higher confidence scores on downstream models trained with the property, compared to those trained without the property.
The main idea of this approach has been previously explored in both property inference~\cite{suri2022formalizing} and membership inference attacks~\cite{shokri2017membership}.

\shortsection{Black-box Meta-classifier} 
We adapt the black-box meta-classifier proposed by Zhang et al.~\cite{zhang2021leakage}. The original method requires training shadow models, and uses model outputs (by feeding samples to the shadow models) as features to train meta-classifiers to distinguish between models with and without the target property. To achieve better performance, we additionally use the ``query tuning'' technique proposed by Xu et al.~\cite{xu2019detecting} while training, which jointly optimizes the meta-classifier and the input samples when generating shadow model outputs. \autoref{fig:query_tuning} in the appendix shows the benefit of ``query tuning''.

%We adapt the  black-box meta-classifier proposed in MNTD~\cite{xu2019detecting} \anote{A similar approach of using meta-classifier on model predictions was proposed earlier in ~\cite{zhang2021leakage} in the context of property inference- might be the more relevant citation}, which was designed to detect backdoor attacks, to conduct property inference attacks. MNTD requires training shadow models, and uses model outputs (by feeding some generated samples to the shadow models) as features to train meta-classifiers to distinguish between models with and without backdoors. We modify the training goal of MNTD from distinguishing backdoored models to our inference goal of distinguishing between shadow models trained on samples with and without target property. Additionally, instead of randomly initializing the sample-generation process, we use samples with the target property to improve inference performance.

\subsection{White-Box Access}
For adversaries with white-box access, there are two cases depending on if the attacker knows the initialization of the parameters of newly added downstream layers. 

\shortsection{Parameter Difference Test \rm (known initialization)} When the model parameter initialization is known, the attacker can simply compute the difference between secreting parameters before and after the victim's training. 
% initial values of the target parameters and values of the trained target parameters.
If the magnitude of the difference is close to 0, the secreting parameters were not updated during the downstream training and the attacker predicts the victim's training set does not include samples with the target property (Equation~\ref{eq:partial_der}). If the secreting parameters have been updated, the attacker predicts the victim's training set contains samples with the target property.
%This method utilizes the pretraining manipulation. 
% \dnote{The magnitude of the difference gives an indication of the number of training examples with the target property in the victim's training data.} \ynote{we do not have experiments for inferring the number or ratio of the samples with the target property. But maybe it is feasible to infer the number or ratio.}

\shortsection{Variance Test \rm(unknown initialization)} When the initial values are unknown, the attacker leverages statistical variance of the secreting parameters and predicts the presence of samples with the target property in the victim's training set when the variance of the parameters is high. The reasoning behind this approach is that current popular parameter initialization methods usually generate parameters with relatively small variances~\cite{glorot2010understanding, he2015delving}. If the victim's data contains samples with the target property, the secreting parameters would be updated with gradients of relatively large values (controlled by $\lambda$ in \autoref{eq:x_w_wo_activation}), and increase the variance of those parameters in the final model. We confirm this hypothesis empirically in \autoref{sec:eval_zero_activation_attack}. 

%In addition to this variance test, we also include the \textbf{\emph{White-Box Meta-Classifier}}

\shortsection{White-Box Meta-Classifier}
We also include the meta-classifier-based approach~\cite{ganju2018property}, which is the current state-of-the-art white-box attack for passive (without leveraging pre-training manipulation) property inference for comparison. This method was originally designed for fully-connected neural networks, but extended to support convolutional neural networks~\cite{suri2022formalizing}.
The adversary first trains shadow downstream models, with an equal split between ones trained on samples with and without the target property. Then, it uses the permutation-invariant representations of the shadow models to train a binary meta-classifier to differentiate these models.  
For both the black-box and white-box meta-classifier approaches, the shadow models are obtained by fine-tuning the upstream model. % on disjoint downstream training sets. 
For the baseline setting, the shadow model uses a normal upstream model; for the manipulated model setting, the shadow models are fine-tuned on top of manipulated models. Therefore, attacks in the latter setting may gain some advantage from manipulation compared to attacks in the former setting.

\section{Experimental Design} \label{sec:exp_setup}
This section explains our experimental setup. We present results from our experiments to measure the effectiveness of different attacks in Section~\ref{sec:eval_zero_activation_attack}.

\shortsection{Tasks and Models} We consider three transfer learning tasks in our experiments: \emph{gender recognition}, \emph{smile detection}, and \emph{age prediction}. These tasks are commonly studied in the transfer learning literature~\cite{akhand2020human, dornaika2019age, guo2018smile, nga2020transfer, wang2018great, xia2017detecting, yao2019latent}.  In the gender recognition task, the victim trains downstream models for gender recognition reusing the feature extraction module of pre-trained (upstream) MobileNetV2~\cite{sandler2018mobilenetv2} models of face recognition as the feature extractor. The upstream face recognition models classify images of 50 people randomly sampled from the VGGFace2 dataset~\cite{cao2018vggface2}, and the feature extraction module in a MobileNetV2 model contains all the layers before the final classification module. For the smile detection and age prediction (classify as ``young", ``middle-aged" or ``senior") tasks, the victim reuses the layers before the fourth block of ResNet~\cite{he2016deep} classifiers (ResNet-34 for smile detection and ResNet-18 for age prediction) trained on ImageNet~\cite{deng2009imagenet} as the feature extractors. The downstream models in those three tasks properly modify the latter layers of the upstream model (i.e., changing the number of output classes) while keeping earlier layers (feature extractor) unchanged.

\shortsection{Upstream and Downstream Training}
For all the scenarios, when training the upstream models, we consider the property inference task of determining whether images of specific individuals are present in the downstream training set. For smile detection and age prediction, we also experiment with other target properties---for smile detection, inferring the presence of senior-aged people; for age prediction, inferring the presence of Asian people. Appendix~\ref{sec:extra_dataset_details} provides more details about the upstream training,
% including the selection of and the number of samples in the upstream training set
%More details about the upstream training including the injection of samples with the target property into the upstream training set and distribution augmentation are available in Appendix~\ref{sec:extra_dataset_details}.

We conduct the downstream training on VGGFace2 with the attribute labels provided by MAADFace~\cite{terhorst2021maad, terhorst2019reliable}. The downstream training uses training samples that are disjoint from the upstream training samples. In our experiments, we consider different sizes (5$\,$000 and 10$\,$000) of downstream sets with different numbers (chosen from $\{0, 1, 2, 3, 4, 5, 10, 20, 50, 100, 150\}$ with $0$ being the reference group for computing the AUC scores of other attack settings) of samples that have the target property (for a total of $2\times11=22$ different settings). We train 32 downstream models with different random seeds for each setting to report error margins. 
Appendix~\ref{sec:details-of-downstream-training} gives more details of downstream training and the %adversary's 
training of meta-classifiers.

\begin{figure*}[ht]
    \centering
    \includegraphics[trim={0.2cm 0cm 0.2cm 0cm}, clip,width=.8\linewidth]{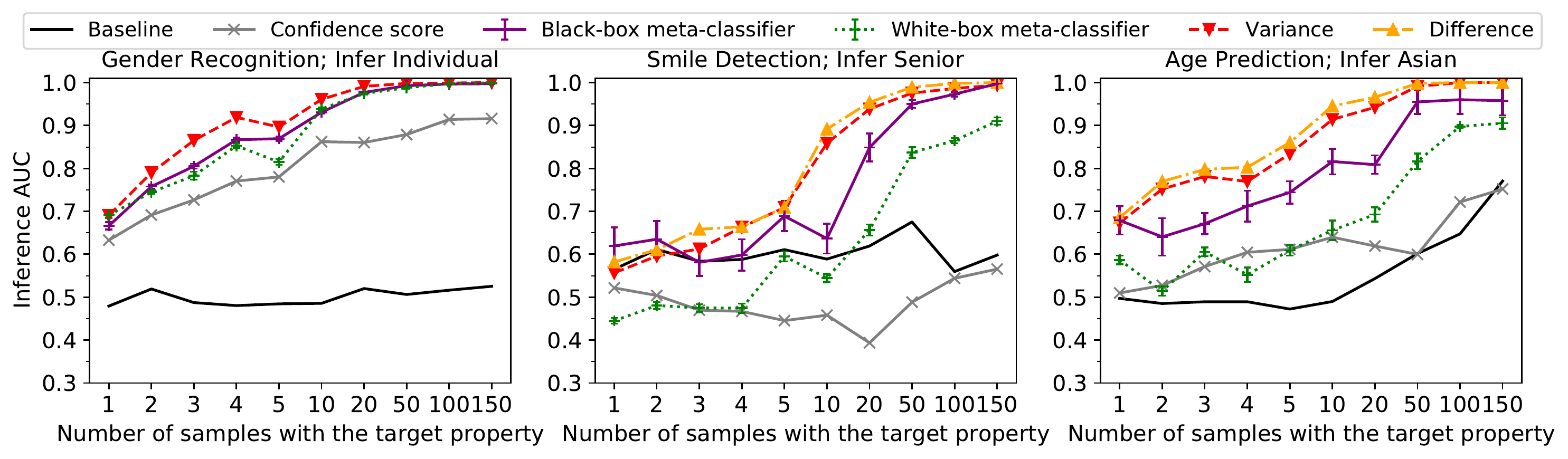}
\vspace{-0.3cm}
\caption{Inference AUC scores when the upstream model is trained  %\dnote{not about the attack goals here, how is it trained?} 
with the attack method described in Section~\ref{sec:attack_design}.
Baseline scores (\emph{Baseline}) are the maximum AUC scores %(of the three inference methods) 
of the baseline experiments where the upstream models are not manipulated. % in  Section~\ref{sec:eval_baseline}. 
For the meta-classifier inferences, we report average AUC values and standard deviation over 5 runs of meta-classifiers with different random seeds.
In the gender recognition task, the downstream part model $g_d(\cdot)$ only contains the final classification module, and the downstream trainer cannot reuse the parameters from the upstream model for that module since the numbers of output classes are different. Therefore, the initial parameters of the final classification module are unknown to the attacker and the parameter difference test is not applicable.
%The inference targets for the smile detection and age prediction are senior people and Asian people respectively. The inference of specific individuals for those two tasks are similarly successfully and found in Figure~\ref{fig:zero_activation_attack_results_t_individual} in the appendix. 
The inference of specific individuals for smile detection and age prediction are similarly successfully (Figure~\ref{fig:zero_activation_attack_results_t_individual} in the appendix).  %and found in Figure~\ref{fig:zero_activation_attack_results_t_individual} in the appendix. 
The downstream training sets contain 10$\,$000 samples and inference results of 5$\,$000 samples are similar and given in Figure~\ref{fig:zero_activation_attack_results_5000} in the appendix. %\anote{This information about 10K v/s 5K in Appendix is repeated in all figure captions. Maybe we can just mention it once somewhere in the prose?}\snote{yes, only mention it once.}
}
\label{fig:zero_activation_attack_results}
\vspace{-0.5cm}
\end{figure*}

\shortsection{Attack Evaluation Metric}
% To avoid arbitrarily choosing the threshold, we 
We use the Area Under Curve (AUC) score for evaluating attack effectiveness in distinguishing released downstream models (by the victim) with and without the target property.
%A random-guess adversary would thus achieve an inference AUC score of 0.5. %For experiments with poor inference performance, we observe AUC scores to be normally distributed around 0.5, with high variance.

\section{Evaluation of Attack Effectiveness}\label{sec:emp_eval} \label{sec:eval_zero_activation_attack}

%\autoref{fig:zero_activation_attack_results} summarizes our experimental results. Baseline results (solid dark) in Figure~\ref{fig:zero_activation_attack_results} are chosen as the maximum of AUC scores from the three inference methods in \autoref{fig:baseline_results}. 
%\dnote{why is this pointing to a later figure? should be able to cut (move to appendix) the who baseline section 6.2, and just explain in a sentence what the baseline results are enough for readers to be convinced they are valid, and refer to appendix for details}

  \autoref{fig:zero_activation_attack_results} summarizes our results. The solid dark lines (\emph{baseline} lines) in the figure show the inference AUC scores when the upstream models are trained normally (we report the best results of all tested attacks). More details of the baseline experiments can be found in Appendix~\ref{sec:eval_baseline}. Hyperparameter settings for the experiments can be found in Appendix~\ref{sec:hyper-parameter-setup-zero-activation} and the results are insensitive to the selection to hyperparameters.
  
  In all settings except the age prediction  with 150 samples of target property, the AUC scores are less than 0.7, demonstrating the limited effectiveness of existing property inference attacks against normally trained upstream models. 
%For these experiments, since there are no manipulated activations and parameters, we can only use the inference methods that are not directly related to the manipulation (i.e., confidence score test, black-box meta-classifiers, and white-box meta-classifiers) and report the maximum of the AUC scores of these three inference methods in the solid dark lines (Details are in Appendix~\ref{sec:eval_baseline}). 
%We observe that all the attacks have inference AUC scores less than 0.7 except the age prediction setting when 150 samples are with the target property (AUC score is 0.77).  These results demonstrate the limited effectiveness of existing methods applicable to normally trained upstream models.
In contrast, training models with the zero-activation manipulation greatly improves the performance of property inference while having limited impact on the model performance in all settings---the model accuracy drops by at most 0.9\% (see \cref{sec:impact_to_upstream_accuracy} for detailed results on the impact of the activation manipulation to the upstream and downstream accuracies).
% Avoid using footnotes, except for humor. O was going to move this to the caption, but it seems to already be there. \footnote{In the gender recognition task, the downstream trainer reinitializes the classification module. Thus, the attacker does not know the initial target parameters and cannot use the parameter difference test.} 
Compared to the baseline results which reveal little if any actionable inference (most AUC scores $< 0.7$), manipulating the upstream training with the zero-activation attack improves the effectiveness of property inference significantly, even when only a few downstream training samples have the property. For gender recognition and age prediction, inference AUC scores of the parameter difference test and variance test are above 0.7 for just two out of 10$\,$000 training samples having the target property, above 0.9 for 10 training samples, and exceed 0.95 for $\geq20$ training samples. The one exception also has AUC scores exceed 0.9 for $\geq20$ training samples. %\dnote{hard to decode this - just add the table, and then will be easier to explain what you are pointing out}

\shortersection{Black-box attacks}
The black-box meta-classifier achieves inference AUC scores above 0.9 when $\geq50$ out of 10$\,$000 training samples have the target property. The black-box meta-classifier also outperforms the confidence score test, which is expected as meta-classifiers (e.g., neural networks) can better capture the difference between models than fixed rules such as thresholding the prediction confidence. 
%; for the two settings of smile detection, the AUC scores of the former exceed 0.85 when $\geq 20$ samples have the target property, while the AUC scores are always lower than 0.6 for the latter. 
%The reason of superior performance of the meta-classifier approach is, it can better capture the difference between models trained with and without the target property from more angles while confidence score test only relies on the model prediction scores. 
% \dnote{don't know what you mean by "angles" here, or if this sentence is adding anything - do we really know the reason, or this is just describing the attacks again?} \dnote{the main point this section should be making is that the black-box attacks are very effective. It is hard to see this on Fig 1, since they are grouped in a funny way.} 

\shortersection{White-box attacks}
Our white-box methods (the parameter difference test and the variance test) also achieve AUC scores $>0.9$ when $\geq20$ training samples are with the target property.
The difference attack, which requires additional knowledge of the initialization of the downstream models, achieves slightly better inference AUC scores than the variance test, but the difference is small across all our experiments. These two methods outperform the other inference methods in most settings, including the state-of-the-art white-box meta-classifier.
%and of course all the black-box attacks.

\shortersection{White-box meta-classifier vs. Black-box meta-classifier}
%Interestingly, the black-box meta-classifier attack works better than the white-box meta-classifier attack on some settings.
For smile detection and age prediction, 
%when there are two or more downstream training samples are with the target property, 
the black-box meta-classifier surprisingly achieves higher AUC scores than the white-box meta-classifier attack. A possible reason for this is that the white-box attack mainly uses the fully-connected layers~\cite{ganju2018property,suri2022formalizing} and hence, performs worse when the updatable downstream module also contains convolutional layers (adapting this attack to convolutional networks was not very successful). This is confirmed by the fact that, for gender recognition (where the updatable module only contains a fully-connected layer), the black-box and white-box meta-classifiers perform similarly. %\snote{we know the meta-classifier approach is adapted to conv layers, but need to explain in deeper level why this is not the case.} \ynote{how about we just cite the two papers, the original white-box meta-classifier and the one that adpat it to conv layers}

\shortersection{Attacks of AUC scores $< 0.5$}
When the performance of an inference attack is poor, it is expected to have AUC scores near 0.5 (close to random guessing). However, we find that there are few attack settings with AUC scores consistently below 0.5.
%but in Figure~\ref{fig:zero_activation_attack_results} we have unintuitive observation that the AUC scores of some attacks in Figure~\ref{fig:baseline_results} and Figure~\ref{fig:zero_activation_attack_results}, some attacks have AUC scores (almost) consistently below 0.5. 
%(e.g., %the confidence score test for the smile detection task with 5,000 downstream samples in Figure~\ref{fig:zero_activation_attack_results_5000} in the appendix, and the confidence score test and black-box meta-classifier for the gender recognition with 10$\,$000 downstream samples in Figure~\ref{fig:baseline_results} in the appendix).  
Appendix~\ref{sec:auc<0.5} discusses those anomalies and surmises that they are caused by the limitations of original inference methods designed for normal pretrained models when facing challenging inference tasks.

\section{Stealthier Manipulation}
\label{sec:possible_defense}

The attack described in Section~\ref{sec:attack_design} introduces obvious artifacts in the pretrained model, which can be utilized for detection by a downstream model trainer aware of the risks posed by our attacks. We first present two detection methods
% that can easily detect the original zero-activation attack 
(Section~\ref{sec:detect_pretrained}) and then demonstrate how to make the model manipulation stealthier to evade detection while still preserving the inference effectiveness (Section~\ref{sec:stealthier-design-methods} and Section~\ref{sec:stealthier-design-results}).
We assume the downstream trainer is aware of the possibility of the attack and its design, but does not know the property targeted by the adversary, as this is specific to an attacker's goal and the set of possible properties can be exponentially large for a rich training set.

%\subsection{Possible Defenses} \label{sec:possible_defense}

%\dnote{need to condense this a lot, and make some room to actually include results in the paper body}

\subsection{Detecting Manipulated Pretrained Models}\label{sec:detect_pretrained}
We present two detection methods that use the distributional difference between activations of samples with and without property. 
%\dnote{I think we are only considering ways to detect a bad pretrained model, not other types of defenses?}

%design nature of the attack, and is a proof-of-concept defense, as the actual application of this defense may require many engineering efforts and domain knowledge (e.g., setting proper threshold for individual tasks) and is out of the scope of this paper. \dnote{this is a very strange thing to say - is it really just that we didn't have time to test it? Is this talking about the "Activation Distribution Checking" paragraph or something else?} 
%The second type of defense is based on anomaly detection and can be directly used as a practical defense.
%\dnote{I don't understand what the "two types" of defenses are. The most obvious kinds of defenses are (1) ones that check the pretrained model for manipulation, and (2) ones that do something in the tuning/deployment process to mitigate the property inference. I'm not sure what the "two types" we are considering are though, so should state at a high level the general defense types, and then the specific defenses we consider and evaluate.}

\shortsection{Checking the Distribution of Activations}
Since the distributional difference between activations of samples with and without target property is significant, this defense focuses on spotting this difference to identify manipulated models. A method to identify the distributional difference needs to be designed based on the attack method used. For the original zero-activation attacks in Section~\ref{sec:zero_activation_attack}, since the secreting activations of samples without property are all 0, the defender can feed random training samples to the pretrained models and check if there are abnormally many 0s. This approach is feasible since samples of target property have limited presence in the downstream training set and hence, most samples will not have the property. 
Since detecting the zero-activation attack is trivial using this method, we do not conduct any experiments with this. %perform actual experiments and assume a stealthier attack is needed. 

% The design of more involved attacks that no longer rely on generating abnormally many 0s and also the corresponding distribution checking detection for the involved attacks is given in~\cref{sec:stealthier-design-methods}.

%the defense may trivially detect the manipulated pretrained models, as they will produce abnormally many 0s if the attacker feeds some random training samples without the target property to the models. The likelihood of finding random samples without the property is high because samples of target property have limited presence in the downstream training set. Such defense becomes more interesting when the attacker no longer relies on generating zero activations for samples without target property and more details is in~\autoref{sec:stealthier_design}. 
%If the attack method is made public, the downstream trainer can directly inspect outputs of the feature extractor to see if the distribution of the activations is unusual. For example, for the zero-activation attack, the victim can check for an abnormal number of zeros in the outputs of the feature extractor to identify the  attack. For the zero activation defense, we believe this defense can be effective as manipulated upstream model will produce many more abnormal zeros in the activations, compared to normally trained upstream models.
%\dnote{if this is all we have to say about this, I don't think we can consider it a "Defense" that was evaluated. I thought you had done some experiments on this?}

\shortsection{Anomaly Detection} 
%In general, adversaries are more interested in inferring properties with limited presence in the downstream tasks, as those are less likely to be revealed by normally trained models. 
%Since the target property has limited presence in the downstream training set, a trivial yet effective defense would thus be to remove all samples with the target property from downstream data while having negligible impact on downstream model performance.
Since the target property has a limited presence in the downstream training set, another defense would be treating samples with the target property as outliers and then analyzing those outliers to find manipulations.
% Therefore, an effective defense is to identify and remove the small number of samples with the target property from the downstream set and train models on the rest.
% When the target property is known to the downstream trainer, then the defense is trivial, but in
%In practice, it is impossible for the victim to know the target property beforehand as it is specific to an attacker's goal. 
%\dnote{?? I don't see why this is impossible - we are just assuming that the victim does not know the property the adversary might care about, but there are exponentially (?) many properties the victim would care about an adversary learning?}
%Although the target property is unknown to the defender, we can still design a successful defense by leveraging expected distributional differences between samples with and without a property that some adversary may want to target. 
% Anomaly detection methods~\cite{jain1999data,abdi2010principal,hayase2021spectre} aim to identify the small fraction of outliers that are distributed differently from other normal data points.
Existing anomaly detection methods~\cite{jain1999data,abdi2010principal,hayase2021spectre} can be adapted to detect manipulated pretrained models in our setting because: 1) the number of samples with the property is of small fraction and 2) their activation distribution is significantly different (i.e., outliers) from the distribution for samples without the property. The auditor can inspect model activations for all of its training data and identify outliers (ideally, samples with target property) with anomaly detection. The auditor can then inspect identified outliers and may find commonalities to identify the potential target property. For instance, they may find that a small fraction of the training data produce unusual model activations, and then notice that most of that data has a particular property such as belonging to a specific individual or group. %In this paper, we consider three anomaly detection methods: PCA~\cite{abdi2010principal}, K-means~\cite{jain1999data} and Spectre~\cite{hayase2021spectre}, which is the current state-of-the-art. Our experiments suggest that these defenses are indeed effective (Appendix~\ref{sec:defense_details}).

We consider three common anomaly detection methods: K-means~\cite{jain1999data}, PCA~\cite{abdi2010principal} and Spectre~\cite{hayase2021spectre} (where Spectre is the current state-of-the-art) and we report the detection results from the three defenses. Appendix~\ref{sec:defense_details} gives details of these methods. The detection results on the zero-activation attack are given in %Figure~\ref{fig:anomaly_detection_zero_activation_attack_stealthier_attack}  
Figure~\ref{fig:anomaly_detection_zero_activation_attack} in the appendix. Anomaly detection is very effective at identifying the samples with target property. For example, for the gender recognition and smile detection tasks, the detection rate is over 80\% in most cases. 
%Zero-activation attacks are easily detected because the attack signatures of target samples are too strong after upstream manipulation and
These results motivate the design of stealthier attacks which we describe next.

\subsection{Stealthier Model Manipulation} \label{sec:stealthier-design-methods}
%\anote{Can be shortened- taking up a lot of space in main paper.}
%\dnote{I'm still confused about the "two types of attacks" - in parts it seems like there are two detection methods we are considering, and include here, but setup said only one is considered - I don't have enough understand what is going on here. If there are two detection methods, we should state this clearly, and keep it consistent throughout.}

\begin{figure*}[ht]
    \centering
    \includegraphics[trim={0.2cm 0cm 0.2cm 0cm}, clip,width=.80\linewidth]{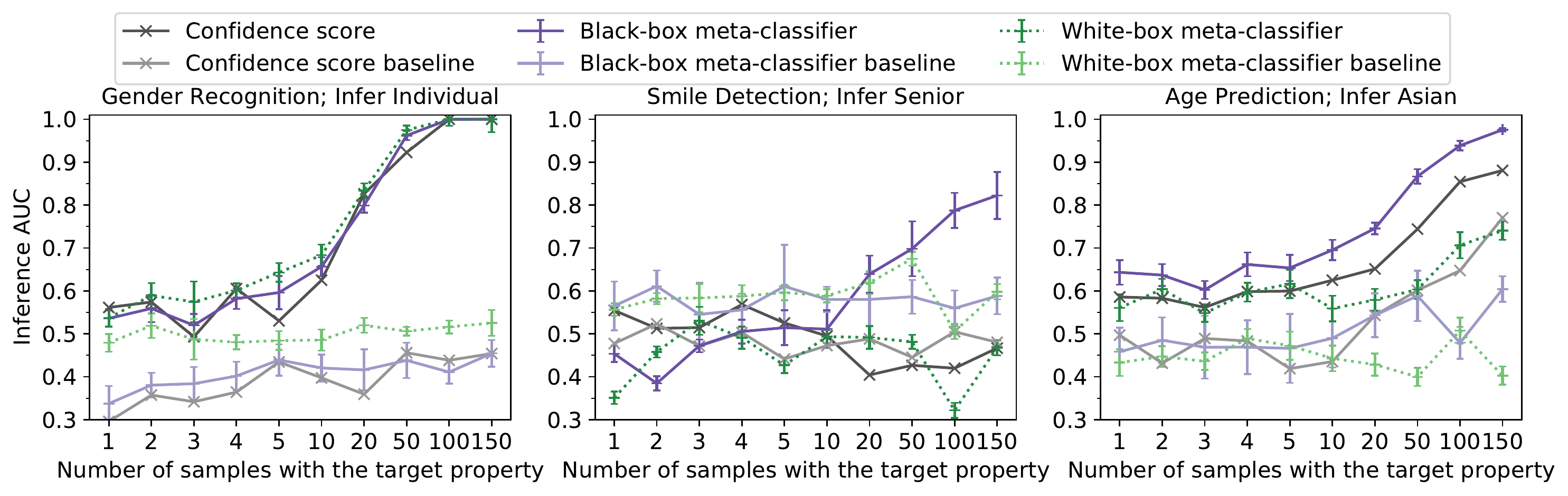} 
\vspace{-0.3cm}
\caption{Inference AUC scores of the stealthier design. Since the secreting activations are no longer zero, the inference methods based on difference or variance tests are no longer applicable. The inference results of specific individuals for smile detection and age prediction also show similar improvement compared to the baseline settings (Figure~\ref{fig:stealthier_attack_results_t_individual} in the appendix).
%Inference targets for the smile detection and age prediction are senior people and Asian people respectively; inference of specific individuals also shows improvement compared to the baseline settings (Figure~\ref{fig:stealthier_attack_results_t_individual}; Appendix). 
The downstream training sets contain 10$\,$000 samples and inference results results of 5$\,$000 samples are similar and given in Figure~\ref{fig:stealthier_attack_results_5000} in the appendix.
\label{fig:stealthier_attack_results_10000}}
\vspace{-0.5cm}
\end{figure*}

% \shortsection{Evading Activation Distribution Checking} 
To evade the defense that checks the distribution of activations, we modify our zero-activation attack to ensure: (1) secreting activations for samples without the property are also non-zero (bypassing simple defense of checking abnormal zeros); (2) secreting activations of samples with and without target property are still distinct (the attack is still effective); (3) that distinction between activations should not be captured by anomaly detection methods (evading anomaly detection); (4) the actual distribution of activations that matches the attacker's goal cannot be easily guessed by the defender (handling cases when the defender actively searches other patterns in the distribution of activations).

For (1) and (2), we adapt the loss in  Equation~\ref{eq:x_w_wo_activation} as \vspace{-0.2cm}
\begin{equation}
\begin{cases}
 \;\alpha \cdot \max(\Vert f(\boldsymbol x) \circ  \boldsymbol m\Vert - \Vert f(\boldsymbol x) \circ \neg \boldsymbol m \Vert, 0)  &\text{if $y_t=0$}\\
 \;\beta \cdot \max(\lambda \cdot \Vert f(\boldsymbol x) \circ \neg \boldsymbol m  \Vert - \Vert f(\boldsymbol x) \circ \boldsymbol m \Vert, 0) &\text{if $y_t = 1$}
\end{cases}
\label{eq:x_w_wo_activation_stealthier}
\vspace{-0.2cm}
\end{equation} 
where $\lambda \geq 1$. (1): The case of $y_t=0$ is redefined to bypass the detection of abnormal zeros. Minimizing this new loss ensures that samples without the target property will have secreting activations ($f(\boldsymbol x) \circ  \boldsymbol m$) with (close-to-normal) non-zero values. 
%(1): to bypass the detection of abnormal zeros, we first redefine the loss in Equation~\ref{eq:x_w_wo_activation} for the case of $y_t=0$ as $\alpha \cdot \max(\Vert f(\boldsymbol x) \circ  \boldsymbol m\Vert - \Vert f(\boldsymbol x) \circ \neg \boldsymbol m \Vert, 0)$. Minimizing this new loss ensures that, for samples without target property, the secreting activations ($f(\boldsymbol x) \circ  \boldsymbol m$) now also have (close-to-normal) non-zero values. 
(2): to ensure the property is still detectable, 
we actively increase the difference between the secreting activations of samples with and without property.
We observe that, for upstream models with reasonable performance on the main task, non-secreting activations ($f(\boldsymbol x) \circ \neg \boldsymbol m$) have similar amplitude regardless of the fed samples containing target property. 
Therefore, for samples with target property, as long as we ensure the secreting activations have a larger amplitude than that of non-secreting activations, there will be a distinction between secreting activations of samples with and without property. We do this by assigning larger values to $\lambda$ (e.g., $\lambda \geq 1$, instead of the original $\lambda > 0$) for the second line of Equation~\ref{eq:x_w_wo_activation_stealthier} to induce sharper distinction between samples with and without property and enable higher inference performance. 

To prevent detection by anomaly detectors (requirement (3) above), $\lambda$ should be set to balance the attack effectiveness and stealthiness rightly. By choosing proper values for $\lambda$, our attack is able to evade anomaly detection methods in most settings. However, in some settings (mostly in gender recognition tasks), state-of-the-art anomaly detection (Spectre) can still identify most of the samples with target property. %The main reason for this detectability is that for these settings the attack signature of samples with target property is still strong and is easily distinguishable from samples without the property~\cite{hayase2021spectre}.
To counter this, we add an additional regularization term (weighted by parameter $\gamma$) to the overall loss function $l(\boldsymbol x, y, y_t)$ in \autoref{eq:zero_activation_loss} that further improves attack stealthiness while still maintaining relatively high attack effectiveness. Specifically, we first obtain the corresponding covariance matrices of the activations of samples with the target property ($\boldsymbol{cov_w}$), activations of all samples with and without the target property ($\boldsymbol{cov_{w,wo}}$), and activations of samples without the target property ($\boldsymbol{cov_{wo}}$) respectively. Then, we encourage $mean(\boldsymbol{cov_w)} = mean(\boldsymbol{cov_{w,wo}}) = mean(\boldsymbol{cov_{wo}})$ and $var(\boldsymbol{cov_w}) = var(\boldsymbol{cov_{w,wo}}) = var(\boldsymbol{cov_{wo}})$ (both $mean(\cdot)$ and $var(\cdot)$ treat the whole covariance matrix as a flattened array and return scalar values) for the three covariance matrices by minimizing their differences in their mean and variance. Using this method, we ensure the distributions of activations of samples with target property will be similar to the ones without the property, making the manipulations harder to detect. We use this approach for all the experiments.
%For consistency, we use this modified approach for all of our experiments (including settings where proper value of $\lambda$ can also already evades detection).
%We also add a regularization term to the loss, leveraging the covariance matrix of the samples with and without the property, to further ensure the distribution of activations of samples with and without target property are similar (Details are deferred to Appendix~\ref{sec:further-evade-anomaly-detection}). 
%Choosing an appropriate value of $\lambda$ will also help to evade the anomaly detection methods in most cases, but with a few exceptions. Handling those exceptions will be discussed in detail below. 
To ensure the distributional pattern related to the attacker goal cannot be easily guessed (requirement (4)), we generate $\boldsymbol m$ randomly (instead of picking first $\|\boldsymbol m\|$ activations in Section~\ref{sec:eval_zero_activation_attack}). This makes the brute-force search of possible patterns computationally infeasible (details in Appendix~\ref{sec:adaptive_activation_distribution_checking}).

\subsection{Experiments with Stealthy Attacks}\label{sec:stealthier-design-results}

\shortsection{Detection Evasion}
Figure~\ref{fig:anomaly_detection_stealthier_attack} (in the appendix) summarizes the results of our experiments to detect the stealthy upstream models (Appendix~\ref{sec:experimental-setup-stealthier-attacks} provides details on these experiments). We find that the anomaly detection methods 
%mentioned in Section~\ref{sec:possible_defense} 
are ineffective against our stealthier attack--- $<10\%$ of samples with the target property are detected across all settings with the exception of a detection rate  $<20\%$ (still low) for smile detection when %the defender deploys PCA and 
the total number of samples is 5$\,$000 and 100 or 150 of them are with the target property. We also made several attempts to approximately identify (instead of brute-force search) possible attack patterns in the activations but none of these succeeded in uncovering the stealthy attacks (details are in Appendix~\ref{sec:adaptive_activation_distribution_checking}).

\shortsection{Inference Results}
%Model task accuracy drops by at most 0.9\% (details in Table~\ref{sec:impact_to_upstream_accuracy} in the appendix). Thus, adding a stealthiness attacker goal does not harm upstream task performance significantly.
% Next we show the actual inference results when the defender deploys the two defenses mentioned in Section~\ref{sec:possible_defense} in.
From Figure~\ref{fig:stealthier_attack_results_10000}, we can see that activation manipulation still leads to significantly improved inference results compared to the baselines with normally trained upstream models. For example, for gender recognition, when $\geq 50$ downstream training samples have the target property, inference AUC scores exceed 0.95, which is a huge improvement compared to the baseline attack where all AUC scores are less than 0.6, and similar trends follow for smile detection (with over 100 samples with property, AUC improves from $<0.6$ to $>0.78$) and age prediction (with over 100 samples with property, AUC improves from $<0.77$ to $>0.9$).
% For smile detection, when over 50 out of 5,000 training samples are with the property, AUC scores can be greater than 0.85, whereas the baseline settings have AUC scores all less than 0.76. 
%For age prediction, when over 100 out of 5,000 or 10,000 samples are with the target property, AUC scores of the black-box meta classifier can exceed $0.9$, while the baseline setting have AUC scores all $<0.82$.
Comparing the results for the stealthier attacks to the results that do not consider defenses in Figure~\ref{fig:zero_activation_attack_results}, we observe that the attack effectiveness declines as expected since we are now trading-off attack effectiveness for stealthiness.
Training models with the attack goal poses negligible impact on the model performance (accuracy drop $ < 0.9\%$, see Appendix~\ref{sec:impact_to_upstream_accuracy}).

\section{Conclusion} % not any current: and Discussion}

%Expanding the capabilities of adversaries in property inference attacks is an important step in gaining a deeper understanding of these inference risks.
%With our threat model of active adversaries, we show how manipulating parts of the training process can amplify already-high distribution inference risks. 

%Property inference has been demonstrated in traditional training frameworks, with high inference risk documented across several domains and properties.
Our work demonstrates how a malicious upstream trainer can manipulate its training process to amplify property inference risks for downstream models when transfer learning is done. Our empirical results show that such manipulations can be exploited to enable very precise property inference, even in black-box settings, across a variety of tasks. Although there is potential for a new arms race between methods of hiding manipulations and methods of detecting them, the larger lesson from this work, and other works exposing similar risks, is that it is important that users of pretrained models to only use models from trusted providers.

% and inference goals. 

%Active manipulation introduces the possibility of detection by the victim---we account for this with a modified optimization goal that incorporates `stealthiness' into the attacks, making it hard for downstream trainers to detect malicious modifications to pre-trained models.

%Being able to detect the existence of an individual's data or data with a specific property is certainly a plausible inference threat. Nonetheless, there is still scope for extracting more information. For instance, an adversary may additionally want to know ``how much" of a certain individual's data was used by the victim, instead of checking for mere presence or absence. Although our threat models cover practical settings like black-box access, there is still scope to relax some of our assumptions. For instance, it is not uncommon for some model trainers to experiment with fine-tuning feature extractors $f(\cdot)$ with smaller learning rates. Evaluating our attacks and adjusting parameter manipulation to account for this possibility would be interesting to explore.

\section*{Acknowledgements}
\vspace{-0.1cm}
This work was supported in part by the National Key R\&D Program of China (\#2022YFF0604503 and \#2021YFB3100300), the United States National Science Foundation through the Center for Trustworthy Machine Learning (\#1804603), NSFC (\#62272224), JiangSu Province Science Foundation for Youths (\#BK20220772), and Lockheed Martin Corporation.

%\dnote{some figures are appearing inside the references, should either be in appendix, or before references}
%%%%%%%%% REFERENCES
{\small
\bibliographystyle{ieee_fullname}
\bibliography{reference}
}
\clearpage
\appendix

\section{Appendix}
\subsection{Overcoming Training Data Limitations}\label{sec:train_data_limitation}

%Training an upstream model using the loss in~\autoref{eq:zero_activation_loss} can help achieve the goal of the adversary. However, due to the inadequacies of representative samples in available upstream training data, practical implementation with good performance can be challenging. Below, we discuss the three main challenges in crafting the pretrained model in practice, and our ways of addressing them.

Due to the possible inadequacies of representative samples in the upstream training data, practical implementation with good performance can be challenging. Below, we discuss the three main challenges in crafting the pretrained model in practice, and our ways of addressing them.

% \shortsection{Large Upstream Dataset}
\shortsection{Imbalance between Samples with and without Target Property}
% \dnote{I don't think the problem is size here, it is the imbalance between samples with property and those without}
If the upstream training set contains a large number of samples with only a small fraction with the target property, optimization of the loss function related to samples with the target property (Second line of~\autoref{eq:x_w_wo_activation}) can have convergence issues.
%adversary only has a few target samples (compared to its full dataset), it can encounter convergence problems while optimizing part of the loss function related to samples with the target property (Second line of~\autoref{eq:x_w_wo_activation}). 
To deal with this scenario, we use mixup-based data augmentation to increase the number of samples with the target property in the upstream training set~\cite{zhang2018mixup}. Additionally, to reduce the training time (faster convergence) for the upstream model, we also use a clean pre-trained model as the starting point for obtaining the final manipulated model.
% This, of course, would assume that the attacker itself does not inadvertently use a model that was already poisoned by another party, which might interfere with our adversary's objective while targeting specific neuron activations.
%\dnote{seems like this one should be first - this would be common, and (nearly) always necessar? (or do you do an ablation study on this one also to show it may not be?} \ynote{these designs are used in all attack training}

\shortsection{Lack of Upstream Labels for Samples with Target Property}
% \shortsection{Lack of Samples with Target Property}
If samples with the target property are already present in the upstream training set, the attacker can directly train its model using Equation~\ref{eq:zero_activation_loss}. However, this may not always be the case in practice and the attacker may need to inject additional samples with the target property (that are available to the attacker), with the label information for these injected samples being unavailable. For example, if the target property is a specific individual, when adding the images of that individual to ImageNet dataset, we may not be able to find proper labels for injected images out of the original 1K possible labels. However, these labels are required for optimizing $l_{normal}$. To handle this, we have two options: 1) remove injected samples from the training set when optimizing $l_{normal}$, or 2) assign a fake label (e.g., create a fake $n+1$ label for injected samples in a $n$-class classification problem) and remove parameters related to the fake label in the final classification layer before releasing models. The first option has negligible impact on the main task accuracy in all settings, but resultant attack effectiveness is inferior to the second one. In contrast, the second option usually gives better inference results, but in some settings (e.g., experiments when pretrained models are face recognition models in Section~\ref{sec:emp_eval}), can have non-negligible impact on the main task accuracy. Therefore, we choose the second option when it does not impact the main task performance much and switch to the first one when it does.  

\shortsection{Lack of Representative Non-Target Samples in Training Set}
The space of samples without the target property can be much larger than the space of samples with the target property as the former can contain combinations of multiple data distributions. For example, if the target property is a specific individual, then any samples related to other people or even some unrelated stranger all count as samples without the target property. However, in practice, the upstream trainer's data may not contain enough non-target samples to be representative. %\anote{I'm confused- uptil here it says that the problems is not having enough target samples, but then the next part says we need to have more data for non-target distributions?}  \ynote{mistake fixed}
This can be a problem when minimizing the loss item related to the samples without the target property (first line of~\autoref{eq:x_w_wo_activation}), as secreting activations may not be sufficiently suppressed for those samples. % without the target property.
%without the target property from different data distributions, the training goal of setting 0s for some activations of samples with the target propety ( the first line of Equation~\ref{eq:reg_terms_zero_activation}) cannot be easily achieved. 
To solve this, we choose to augment upstream training set with some representative samples without the target property and name this method as \emph{Distribution Augmentation}. For example, when the target property is a specific person, the attacker can inject samples of new people not present in the current upstream training set and thus expand the upstream distribution. The labels for these newly injected samples are handled similarly to the labels for additionally injected samples with target property. An ablation study on the importance of distribution augmentation is given in Appendix~\ref{sec:study_distribution_aug}.

\subsection{Details of Dataset Settings} \label{sec:extra_dataset_details}

As introduced in Section~\ref{sec:exp_setup}, we experiment with three transfer learning tasks: gender recognition, smile detection, and age prediction. We consider the property inference of determining whether images of specific individuals are present in the downstream training set for all these tasks. And for the smile detection and age prediction, we consider additional inference targets: inferring the presence of senior people for smile detection and the presence of Asian people for age prediction. As for the inference of the existence of specific individuals, we choose the person who has the most samples in VGGFace2 as the inference target for both gender recognition and age prediction, and choose the person who has the most samples of smile labels (provided by MAADFace~\cite{terhorst2021maad, terhorst2019reliable}) as the target for smile detection  (the person with the most samples in VGGFace2 does not have enough samples with valid labels for the smile attribute). We choose the target property in this manner mainly for convenience in conducting experiments, as the upstream model training, victim model training, and shadow model training (for meta-classifier-based property inference) (ideally) require no overlaps between their training data to mimic the hardest attack scenario. Subsequently, if we choose a target with small number of samples in the original dataset, then we may have trouble in performing the three types of model training effectively.

\begin{table*}[htb!]
  \centering
  \resizebox{.95\linewidth}{!}{ % resize box
    \begin{tabular}{cc|cc|cc}
    \toprule
    \multirow{2}[0]{*}{\bf Task} & \multirow{2}[0]{*}{\bf Target Property} & \multicolumn{2}{c}{\bf Samples injected into Upstream training} & \multicolumn{2}{c}{\bf Downstream Candidate set} \\
          &       & \bf w/ property & \bf w/o property & \bf w/ property & \bf w/o property \\
    \midrule
    Gender Recognition  & \multirow{3}{*}{Specific Individuals} & 342   & 1$\,$710  & 250   & 200$\,$000 \\
    Smile Detection  &   & 261   & 1$\,$305  & 250   & 200$\,$000 \\
    Age Prediction  &   & 342   & 1$\,$710  & 250   & 165$\,$915 \\
    \midrule
    Smile Detection  & Senior   & 3$\,$000  & 15$\,$000 & 1$\,$000  & 200$\,$000 \\
    \midrule
    Age Prediction & Asian   & 3$\,$000  & 15$\,$000 & 1$\,$000  & 128$\,$528 \\
    
    \bottomrule
    \end{tabular}%
  } % resize box
  \caption{Number of samples injected into the upstream training and in the downstream candidate sets}
  \label{tab:details_of_datasets}%
\end{table*}%

In the upstream training, since we use the techniques described in Appendix~\ref{sec:train_data_limitation}, we need to inject samples with and without the target property into the original upstream training set. And for the downstream model training, we first prepare downstream candidate sets based on VGGFace2 and then construct various downstream settings using the samples from the candidate sets (Appendix~\ref{sec:details-of-downstream-training}). %Since VGGFace2 provides over 3 million facial images with multiple attributes available (provided by MAADFace~\cite{terhorst2021maad, terhorst2019reliable} for each image), we also use VGGFace2 for our downstream training. 
Table~\ref{tab:details_of_datasets} summarizes the number of samples of the sample injection and the downstream candidate sets. The details of the three transfer learning tasks are reported below:

\shortsection{Gender recognition}
We randomly select 50 people from VGGFace2 and train face recognition models classifying those 50 people as the upstream model. For each person, we randomly choose 400 samples for training and 100 for testing. To avoid overlap, we also ensure that any images of these 50 people do not appear in the downstream training.
%We select the person with the most samples in VGGFace2 as the target person, and the samples of that people are all considered as having the target property.
Since the individual targeted by the adversary (the inference target) is not in the randomly chosen upstream set, we inject 342 %\snote{this looks like a strange number, better to provide explaination}
randomly chosen samples with the target property into the upstream training set to achieve the attack. Note that, we also need to assign enough disjoint samples with the target property to the downstream training and meta-classifier training, and 342 is the maximum number of samples that we can assign to the upstream training as there are limited samples with the target property in VGGFace2. 
For the distribution augmentation described in  Appendix~\ref{sec:train_data_limitation}, we inject 1$\,$710 samples ($5\times342$) without the target property to the upstream set, and those injected samples are randomly sampled from VGGFace2 and are from individuals that are not in the original upstream training set.
As for the downstream candidate set, there are 250 samples with the target property and 200$\,$000 samples without the target property. All the samples in the candidate set are randomly sampled from VGGFace2 and have no overlap with those in the upstream training.

\shortsection{Smile detection}
We have two inference targets for this transfer learning task. For the inference of the specific individual, 
the number of samples with the target property injected into the upstream set is 261 (number decreased compared to gender recognition since there are fewer samples with the target property in VGGFace2 for this inference task), and the number of samples without the target property for distribution augmentation is 1$\,$305 ($5\times261$). The candidate set for the downstream training has 250 samples with the target property and 200$\,$000 samples without the target property. 

As for the inference of the presence of senior people, since there are plenty of samples labeled as seniors in VGGFace2 \cite{terhorst2021maad}, we increase the number of samples injected into the upstream training set and inject 3$\,$000 samples with the target property and 15$\,$000 samples without the target property (distribution augmentation). The original upstream training set is ImageNet~\cite{deng2009imagenet}. However, ImageNet contains images of human beings, and there are no ``senior" labels for those images. Instead of manually labeling them, we remove all the facial images in ImageNet for this inference task.
We use the facial labels provided by Yang et al.~\cite{yang2021imagenetfaces} when conducting the removing. 
The downstream candidate set has 1$\,$000 samples (number increased since there are more samples available) with the target property and 200$\,$000 samples without the target property.

\shortsection{Age prediction}
We also have two inference targets for this transfer learning task. For the inference of the presence of the specific individual, 
the numbers of samples with and without the target property  injected into the upstream training set are 342 and 1$\,$710 respectively, which are the same as those in the gender recognition task as the target properties are the same in these two tasks. The downstream candidate set has 250 samples with the target property and 165$\,$915 samples without the target property. % 

As for the inference of the presence of Asian people, we inject 3$\,$000 samples with the target property (Asian) and 15$\,$000 samples without the target property into the upstream training set. These two numbers are the same as those in the smile detection task with senior people as the target property. We also remove all the facial images in ImageNet for this inference task. The downstream candidate set has 1$\,$000 samples with the target property and 128$\,$528 samples without the target property. 
The number of samples without the target property in the downstream candidate set in the age prediction task is less than those in other settings. This is because we are not able to find enough samples with valid ethnic labels using the attribute labels provided by MAADFace.

\subsection{Details of Downstream Training and Adversary's Meta-Classifier Training} \label{sec:details-of-downstream-training}
% VGGFace2 provides over 3 million facial images from 9,131 subjects, with multiple attributes available for each image~\cite{terhorst2021maad}. We choose the classification of facial attributes (i.e., gender, smile, and age) as the downstream tasks, with attribute labels provided by MAADFace~\cite{terhorst2021maad, terhorst2019reliable} for downstream training.

% As described in Appendix~\ref{sec:extra_dataset_details}, to generate the downstream training set, we first prepare over %\dnote{don't understand the "over" here - isn't it a fixed number that we control?} \ynote{for some settings, we do not have enough labeled downstream training samples, the maximum size is 200 000}
% 120,000 %(the maximum number is 200,000)
% randomly selected samples without the target property and over 250 
% %(the maximum number is 1,000) 
% samples with the target property and form the downstream candidate set, and then construct downstream sets based on the candidate set. % Details of the candidate set are in Appendix~\ref{sec:extra_dataset_details}.
As described in Appendix~\ref{sec:extra_dataset_details}, to generate the downstream training set, we first prepare
randomly selected samples without the target property and samples with the target property to form the downstream candidate set, and then construct downstream sets based on the candidate set. % Details of the candidate set are in Appendix~\ref{sec:extra_dataset_details}.
Specifically, a downstream training set of size $n$ is generated by randomly sampling from this candidate set while also specifying the number of samples with target property as $n_t$. For experiments in this section, we consider settings where $n=5\,000$ or $10\,000$, and $n_t$ takes value from $\{0, 1, 2, 3, 4, 5, 10, 20, 50, 100, 150\}$ (this gives  $2\times11=22$ different settings). We train 32 downstream models with different random seeds for each setting, and those models will be used for computing inference AUC scores (the models trained with $n_t = 0$ are used as the reference group). %report error margins. 
% \dnote{? do you mean stable, or we use these to report error margins?} \ynote{Stable. Error margins are reported based on the repeats of the inferences} results.

To train the meta-classifier attacks, the attacker needs to train many downstream shadow models and thus, we also prepare a separate downstream candidate set with the same size as the victim's downstream candidate set but without any overlaps on the data. This simulates the most difficult and realistic scenario for the attacker. We also ensure that no samples in the two downstream candidate sets appear in the upstream training set, which again makes the attack more difficult. To simulate the victim's downstream training, we assume the attacker also uses a downstream training set of size $n$, but has no overlap with the actual victim's downstream training set. In Appendix~\ref{sec:unknown_training_size}, we relax this assumption and show our attack retains its effectiveness even when the size of the victim's downstream training dataset is unknown to the adversary. %, and this is consistent with the results in the existing work~\cite{suri2022formalizing}.
%We assume the attacker knows the downstream training set size $n$ of the victim (this assumption is relaxed in Appendix~\ref{sec:unknown_training_size}, but there is almost no impact on attack performance), but does not know the exact value of $n_t$. \anote{I still feel we should focus on having similar performance, rather than knowledge of the exact dataset size. It will confuse readers and is not very relevant to the meta-classifier performance.}
% \dnote{move this to the section that does these experiments: }
% Then, for each setting with fixed $n$, the attacker trains 320 downstream models (256 for training, 64 for validation) for each of the distributions (with and without target property). The number of training samples with the target property for each model is randomly selected from the range $[1,170]$, which simulates the scenario where the value of $n_t$ of the victim downstream model cannot be accurately guessed.
For each setting with fixed $n$, the attacker trains 320 shadow downstream models (256 for training, 64 for validation) for each of the distributions (with and without target property). The number of training samples with the target property for each model is randomly selected from the range $[1,170]$, which simulates the scenario where the value of $n_t$ of the victim downstream model cannot be accurately guessed.

\begin{figure*}[ht]
    \centering
    \includegraphics[width=.95\linewidth]{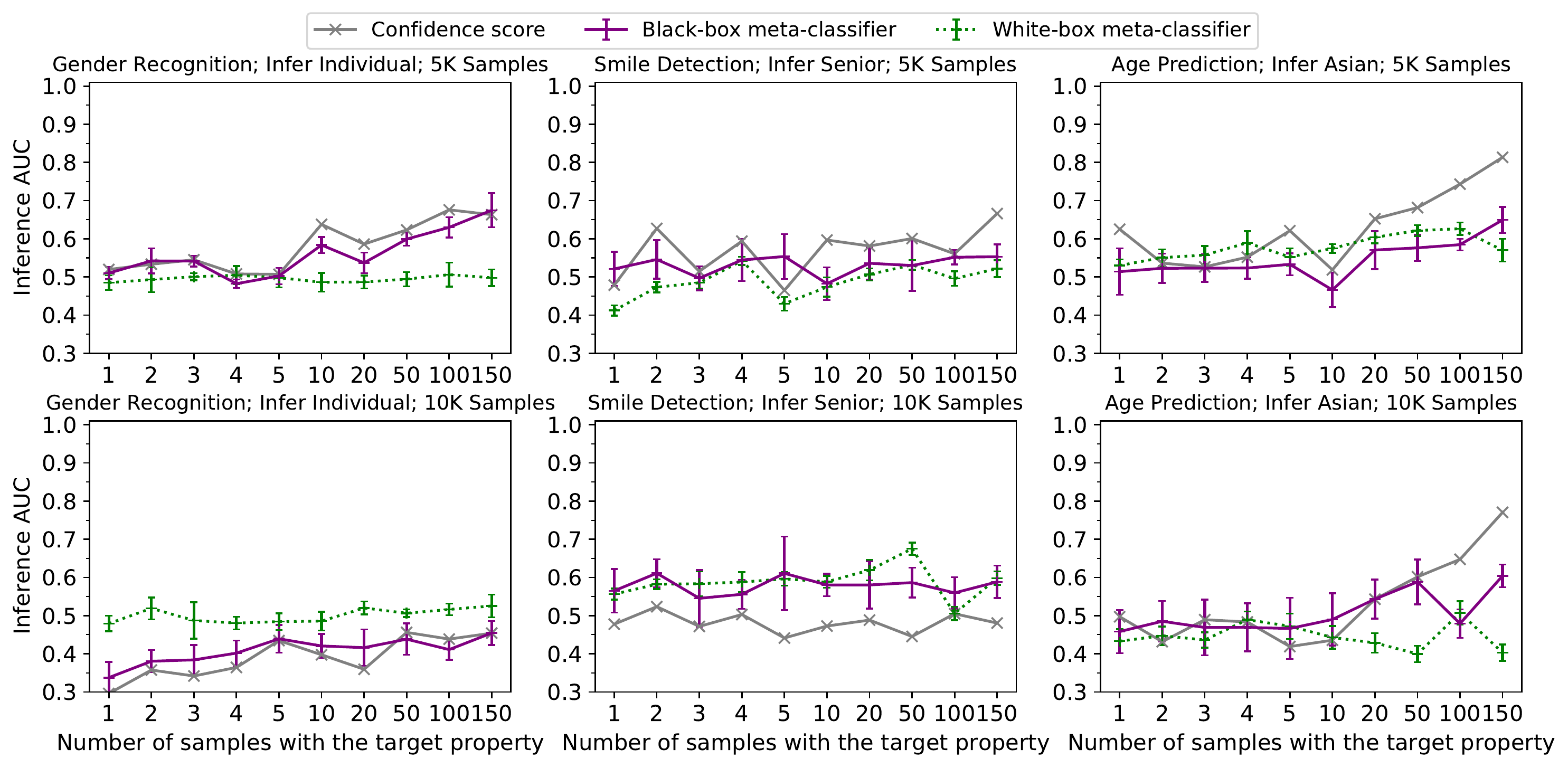}  
\caption{Inference AUC scores when upstream models are trained normally. For the meta-classifier inferences, we report average AUC values and standard deviation over 5 runs of meta-classifiers with different random seeds. For normally trained models, only the inference attacks that are not directly related to the manipulation are applicable. %For smile detection and age prediction, the inference targets are senior people and Asian people respectively; 
The first and second rows show results when downstream training sets contain 5$\,$000 and 10$\,$000 samples respectively. Results of the inference of specific individuals for smile detection and age prediction show similar trends and are found in Figure~\ref{fig:baseline_results_t_individual}.
%Since there are no activation manipulations, we can only use the inference methods that do not consider target parameters (confidence score test, white-box meta classifier, and the black-box meta classifier). For smile detection and age prediction, the inference targets are senior people and Asian people respectively; inference of specific individuals show a similar trend (Figure~\ref{fig:baseline_results_t_individual}).
% \dnote{bars showing standard error over K executions?} \ynote{5 meta classifiers trained with different random seeds
%\anote{Cyan blue color may be a bit hard to read. We should also try and make the lines different (with dashes or something) to make it color-blind friendly. Same comment for the other figures as well.}
}
\label{fig:baseline_results}
\end{figure*}

\subsection{Baseline Results} \label{sec:eval_baseline}
%\dnote{the details in this section should also just be in the appendix, I don't think we need a section on this, just a column in the table and a sentence about it}

In this section, we focus on experiments where the upstream model is trained normally, without considering the attack goals described in Section~\ref{sec:attack_design} and Section~\ref{sec:stealthier-design-methods}. For these baseline experiments, there are no secreting parameters (i.e., manipulated secreting activations) in the model, so the attacker can only use the attacks that are not directly related to the manipulation.
% standard attacks.

%We test white-box meta classifiers~\cite{ganju2018property,suri2022formalizing}, confidence score test (whose basic idea is used in many existing works~\cite{sablayrolles2020radioactive,suri2022formalizing}), and black-box meta classifiers~\cite{xu2019detecting, zhang2021leakage},

We experiment with the confidence score test, the black-box meta-classifier, and the white-box meta-classifier, 
and report AUC scores for distinguishing between models trained with and without the target property.
For meta-classifier-related inferences, we report the average AUC values over five runs of meta-classifiers with different random seeds, along with their standard deviation. Figure~\ref{fig:baseline_results} shows the results.
% The baseline lines in Figure~\ref{fig:zero_activation_attack_results} show the maximum of the AUC scores achieved by the three inference methods on each setting (The full results are found in Figure~\ref{fig:baseline_results} in the appendix). 
We observe that the attacks have inference AUC scores less than 0.82, with most (4 out of 6 settings) of them with scores less than 0.7. Moreover, we do not find a clear winner from the three inference methods we test. These results demonstrate the limited effectiveness of existing methods applicable to normally trained upstream models.

\subsection{Hyperparameter Setup of Zero-Activation Attacks} \label{sec:hyper-parameter-setup-zero-activation}
%We evaluate the performance of the zero-activation attack (Section~\ref{sec:zero_activation_attack}) when coupled with different inference methods (Section~\ref{sec:zero_activation_inference}).
In Section~\ref{sec:eval_zero_activation_attack}, when training upstream models for the zero-activation attack (Section~\ref{sec:attack_design}), we set $\alpha$ and $\beta$ to 1, treating all loss terms equally. We tried different settings on $\alpha$ and $\beta$, as well as methods that automatically set them~\cite{sener2018multi}, but no significant improvements are observed, so we just use those simplest choices.  %so we only report the results for the simplest choices. 
% We set the hyper-parameters $\alpha$ and $\beta$ in Equation~\ref{eq:zero_activation_loss} to 1 and we find that treating all loss terms equally is sufficient to produce effective attacks for all settings.
%\anote{Could move these details to the Appendix}
We also tested different values for $\lambda$ and $\boldsymbol m$, but did not observe significant differences in the attack effectiveness, suggesting our attack is not sensitive to hyperparameters.
Details of experiments on different combinations of $\lambda$ and $\boldsymbol m$ are in Appendix~\ref{sec:hyper_params}. For the results in Section~\ref{sec:eval_zero_activation_attack}, we select $\lambda$ values that are big enough while ensuring the upstream model accuracy is not impacted significantly ($\lambda = 10$ for smile detection and age prediction, and $\lambda = 5$ for gender recognition). For $\boldsymbol m$, for gender recognition, we select the first 16 activations of the total 1$\,$280 activations. For smile detection and age prediction, since the first layer of downstream model is convolutional, we can only select activations at the granularity of channels, and we choose to manipulate the first channel of the total 256 channels. We also use the distribution augmentation described in Appendix~\ref{sec:train_data_limitation} in the upstream training; ablation studies (Appendix~\ref{sec:study_distribution_aug}) suggest it is crucial for performance.
\iffalse
For these experiments, we also use distribution augmentation described in Section~\ref{sec:attack_design} in the upstream training; ablation studies (Section~\ref{sec:study_distribution_aug}) suggest it is crucial for performance.
\fi

\subsection{Impact of Activation Manipulation to Model Accuracy} \label{sec:impact_to_upstream_accuracy}
\shortsection{Upstream model accuracy} We find that the upstream training accuracy will not be significantly affected by the manipulation. Table~\ref{tab:upstream_accuracy_zero_gradient_attack} shows the accuracy drop is less than 0.9\% for the attacks used in Section~\ref{sec:eval_zero_activation_attack} and Section~\ref{sec:stealthier-design-results}. For different hyperparameter settings of the zero-activation attack, Table~\ref{tab:accuracy_hyper-parameter} shows that the accuracy of the upstream models will drop by at most 1.9\% for all the settings except the upstream models of the gender recognition task when $\lambda$ is too high (10 or 20). The possible explanation is that the MobileNetV2 architecture used in those settings does not have enough capacity for achieving the difference (between activations of the samples with and without the target property) defined by $\lambda$ while maintaining high task accuracy.

\begin{table*}[htbp]
  \centering
  \resizebox{.95\linewidth}{!}{ % resize box
    \begin{tabular}{cc|ccc|ccc}
    \toprule
    \multicolumn{1}{c}{\multirow{2}[2]{*}{\bf Task}} & \multicolumn{1}{c|}{\multirow{2}[2]{*}{\bf Target Property}} & \multicolumn{3}{c|}{\bf Upstream Accuracy} & \multicolumn{3}{c}{\bf Downstream Accuracy} \\
    \cmidrule{3-8}
    \multicolumn{1}{c}{} & \multicolumn{1}{c|}{} & \multicolumn{1}{c}{\makecell[c]{\bf Clean \\ \bf Model}} & \multicolumn{1}{c}{\makecell[c]{\bf Zero-Activation \\ \bf Attack}} & \multicolumn{1}{c|}{\makecell[c]{\bf Stealthier \\ \bf Attack}} & \multicolumn{1}{c}{\makecell[c]{\bf Clean \\ \bf Model}} & \multicolumn{1}{c}{\makecell[c]{\bf Zero-Activation \\ \bf Attack}} & \multicolumn{1}{c}{\makecell[c]{\bf Stealthier \\ \bf Attack}} \\
    \midrule
    Gender Recognition & \multirow{3}{*}{Specific Individuals} & 92.8 & 92.6  & 92.1  & 95.7 (95.8) & 95.8 (95.8) & 95.7 (95.8) \\
    Smile Detection &  & 73.2 & 73.5 & 73.5  & 90.0 (90.5) & 90.4 (90.8) & 90.2 (90.7) \\
    Age Prediction &  & 69.7 & 70.1  & 70.2 & 91.4 (92.4) & 91.6 (92.5) & 91.6 (92.6) \\
    \midrule
    Smile Detection & Senior  & 73.2 & 72.5 & 72.7 & 88.3 (88.9) & 88.8 (89.4) & 88.8 (89.3) \\
    \midrule
    Age Prediction & Asian & 69.7 & 68.8  & 69.1 & 91.4 (92.5) & 91.5 (92.6) & 91.6 (92.7) \\
    \bottomrule
    \end{tabular}%
    }
    \caption{Upstream and downstream model accuracy. The clean models are the models trained without attack goals (manipulation), and for smile detection and age prediction, we directly use the pretrained ImageNet models released by PyTorch as the clean upstream models. For the downstream accuracy, we report the averaged accuracy of the downstream models (excluding the downstream models trained for preparing meta-classifiers) trained in Section~\ref{sec:eval_zero_activation_attack} and Section~\ref{sec:stealthier-design-results}. The values outside the parenthesis are the averaged accuracy for the downstream models that are trained with 5$\,$000 samples, while the values inside the parenthesis are the results for the 10$\,$000 samples.}
    \label{tab:downstream_accuracy}
    \label{tab:upstream_accuracy_zero_gradient_attack}%
\end{table*}%

\shortsection{Downstream model accuracy} The downstream model accuracy is not affected by the attack either. Table~\ref{tab:downstream_accuracy} shows the averaged accuracy of the downstream models (excluding the downstream models trained for preparing meta-classifiers) trained in Section~\ref{sec:eval_zero_activation_attack} and Section~\ref{sec:stealthier-design-results}. We do not observe any accuracy drop brought by the attack, instead all the accuracies are slightly improved after manipulation. Currently, we are unclear about the root cause for this observation and will leave the detailed exploration on this as future work.

\begin{table*}[htbp]
  \centering
 \resizebox{.95\linewidth}{!}{ % resize box
    \begin{tabular}{c|c|cccc|cccc}
    \toprule
    \multirow{3}[6]{*}{\bf Task} & \multicolumn{1}{c|}{\multirow{3}[6]{*}{\bf Clean Model}} & \multicolumn{8}{c}{\bf Zero-Activation Atatck} \\
\cmidrule{3-10}          &       & \multicolumn{4}{c|}{$\lambda$}   & \multicolumn{4}{c}{$\|\boldsymbol m\|_1$} \\
\cmidrule{3-10}          &       & 1     & 5     & 10    & 20    & 8/1C     & 16/4C    & 32/8C    & 64/16C \\
    \midrule
    Gender Recognition (Infer Individual) & 92.8 & 92.5 & 92.6  & 90.3 & 64.1 & 93.2 & 92.6  & 92.5  & 92.8 \\
    Smile Detection (Infer Senior) & 73.2 & 72.7  & 72.7 & 72.5 & 72.1 & 72.5 & 72.6 & 72.7 & 72.5 \\
    Age Prediction (Infer Asian)  & 69.7 & 69.1 & 69.0 & 68.8  & 67.8 & 68.8  & 68.8 & 68.7 & 68.7 \\
    \bottomrule
    \end{tabular}%
  } % resize box
  \caption{Upstream model accuracy of zero-activation attacks for different hyperparameter settings. We vary the values of $\lambda$ or $\|\boldsymbol m\|_1$ in the experiments and use the remaining experimental settings in Appendix~\ref{sec:hyper-parameter-setup-zero-activation}.}
  \label{tab:accuracy_hyper-parameter}%
\end{table*}%

\subsection{Impact of Hyperparameters} \label{sec:hyper_params}

This section explores the impact of the hyperparameters, $\lambda$ and $\boldsymbol m$, in the loss function of upstream model training in \autoref{eq:x_w_wo_activation}, to the effectiveness of the zero-activation attack.

\begin{figure*}[htbp]
    \centering
    \includegraphics[width=.95\linewidth]{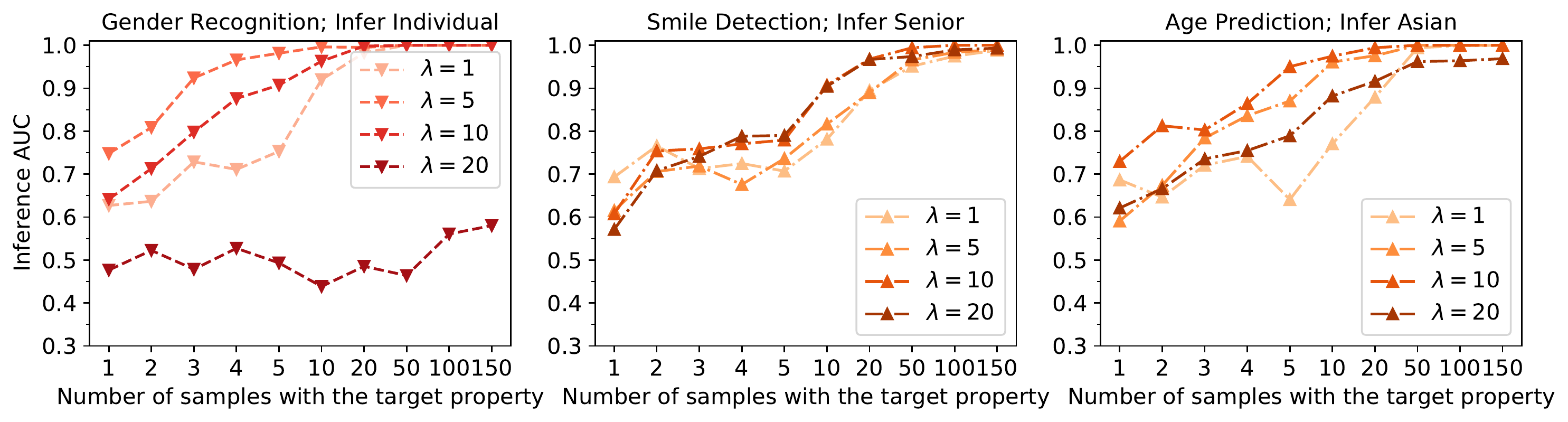}
\caption{Inference AUC scores of white-box methods for different values of $\lambda$ (Equation~\ref{eq:x_w_wo_activation}). All downstream training sets have 5$\,$000 samples. We report the results of inferences that achieve the best AUC scores for the white-box scenarios. Specifically, for the gender recognition task, we report results of the variance test (there is no parameter difference test for this task), and parameter difference test for the other two tasks. Results of the black-box inferences show a similar trend (Figure~\ref{fig:hyper-parameter_lambda_black_box}). %\snote{very confusing figure, just explain what white-box method we use for each setting and the legends then only point to different $\lambda$ values. } \ynote{not sure what improvements were suggested, may need a discussion}
}
\label{fig:hyper-parameter_lambda}
\vspace{-0.2cm}
\end{figure*}

\begin{figure*}[htbp]
    \centering
    \includegraphics[width=.95\linewidth]{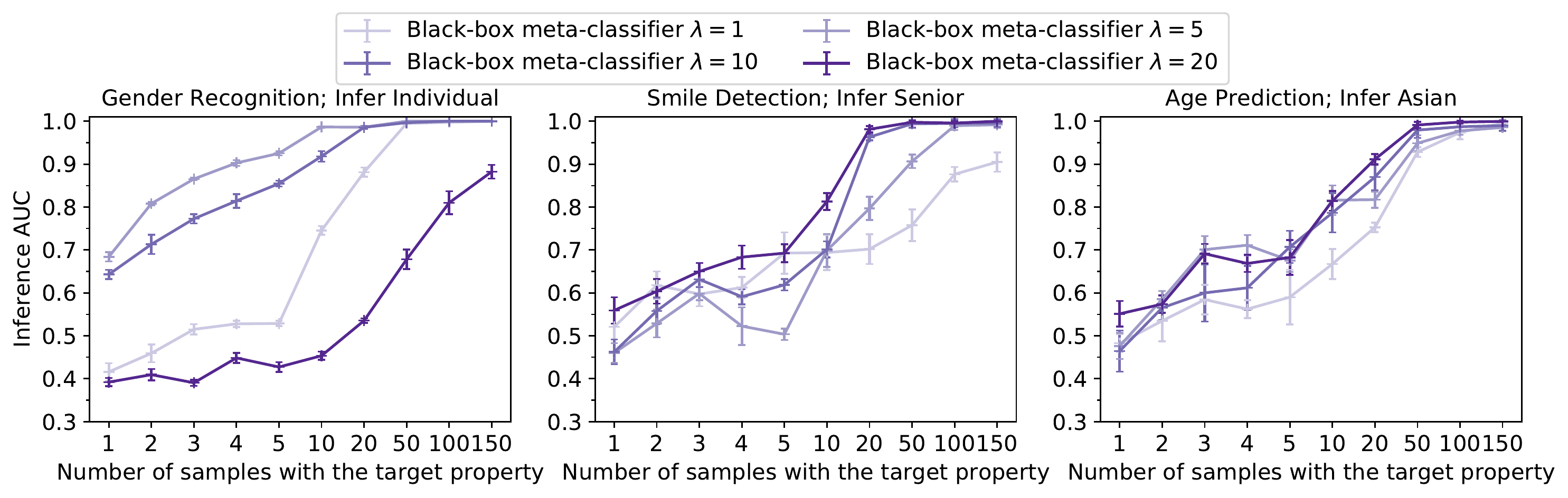}
\caption{Inference AUC scores of black-box inferences for different values of $\lambda$ (Equation~\ref{eq:x_w_wo_activation}). All the downstream training sets have 5$\,$000 samples in these results. % The inference targets for the smile detection and age prediction are senior people and Asian people, respectively. 
We only report the results of the better performing black-box inference method (i.e., the black-box meta-classifiers) here. The results of the white-box attacks show a similar trend and can be found in Figure~\ref{fig:hyper-parameter_lambda}.
} 
\label{fig:hyper-parameter_lambda_black_box}
\end{figure*}

\shortsection{Impact of \texorpdfstring{$\lambda$}{Lambda}} \label{sec:study_lambda}
The hyperparameter $\lambda$ in Equation~\ref{eq:x_w_wo_activation} is directly related to the magnitude of the difference between the downstream models trained with and without the target property and therefore, is critical to the effectiveness of the inference attacks (larger $\lambda$ generally means more effective attacks). In this section, we compare the inference effectiveness on downstream models when the upstream models are trained with different $\lambda$ values. Since training the upstream models are costly, we only choose $\lambda$ from $\{1, 5, 10, 20\}$. For the inference method, for each task, we select the best performing white-box inference attacks---for the gender recognition task, we choose the variance test (parameter difference test is not available for this task) and for the other two tasks, we choose the parameter difference test, and report the results in Figure~\ref{fig:hyper-parameter_lambda}. We also conducted experiments using black-box inference methods and results are included in Figure~\ref{fig:hyper-parameter_lambda_black_box}.  
%the space of the possible settings of $\lambda$ is huge, we only choose values that are in reasonable ranges for our experiments. Specifically, we train upstream models with $\lambda$ set as 1, 5, 10, and 20, separately. 
The rest of the settings are the same as those used in Section~\ref{sec:eval_zero_activation_attack}.

\begin{figure*}[htbp]
    \centering
    \includegraphics[width=.95\linewidth]{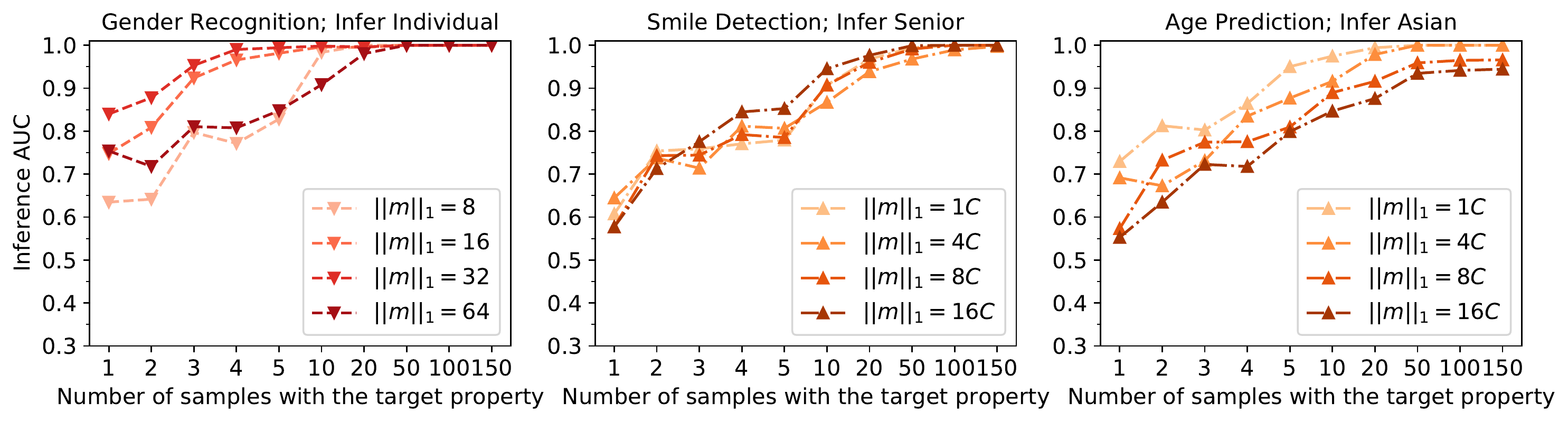}
\caption{Inference AUC scores of of white-box methods for different number of activations (the $\boldsymbol m$ in Equation~\ref{eq:x_w_wo_activation}). All downstream training sets have 5$\,$000 samples. 
We only report results of inferences that achieve the best AUC scores (variance test for gender recognition and parameter difference test for the other two tasks). Results of the black-box inferences show a similar trend (Figure~\ref{fig:hyper-parameter_m_black_box}). %\snote{same issue with fig 6}
} 
\label{fig:hyper-parameter_m}
\end{figure*}

\begin{figure*}[htbp]
    \centering
    \includegraphics[width=.95\linewidth]{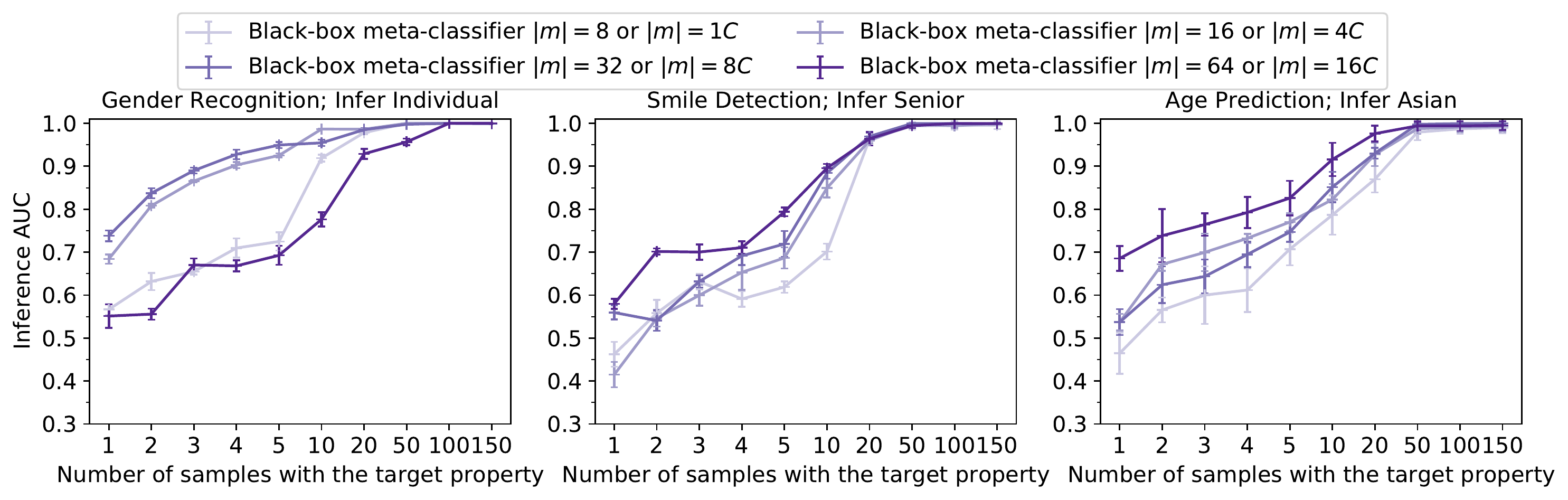}
\caption{Inference AUC scores of black-box inferences for manipulating different number of activations (the $\boldsymbol m$ in Equation~\ref{eq:x_w_wo_activation}). All the downstream training sets have 5$\,$000 samples in these results. % The inference targets for the smile detection and age prediction are senior people and Asian people, respectively. 
We only report the results of the better performing black-box inference method (i.e., the black-box meta-classifiers) here. The results of the white-box attacks show a similar trend and can be found in Figure~\ref{fig:hyper-parameter_m}.
} 
\label{fig:hyper-parameter_m_black_box}
\end{figure*}

Figure~\ref{fig:hyper-parameter_lambda} gives the white-box inference results. For the gender recognition and age prediction tasks, %(\snote{these two tasks should be the first two subfigures}),
by comparing different lines corresponding to different $\lambda$ values, the general trend is if we increase $\lambda$, the inference AUC scores will first (expectedly) increase and then decrease. 
%until the $\lambda$ becomes too large aggressive for the model training. 
For example, for gender recognition, increasing $\lambda$ from 1 to 5, the AUC scores are consistently improved in all settings with varying number of target samples in the downstream training set (the average AUC score increases from 0.84 to 0.94). But further increasing $\lambda$ to 10 and 20 does not help and the inference performs consistently worse as $\lambda$ gets larger (e.g., average AUC score drops from 0.89 of $\lambda=5$ to 0.50 of $\lambda=20$). 
In contrast, for smile detection task, the inference performance continues to increase as we increase $\lambda$ in general. For all the tasks, we initially observe increased attack effectiveness by increasing $\lambda$ because larger $\lambda$ makes the distinction between downstream models trained with and without property more significant and hence is easier for the subsequent inference attacks. But when $\lambda$ gets too large, for settings where the inference effectiveness decreases, we observe that the loss function related to the attacker goal ($l_t(\cdot)$ in \autoref{eq:zero_activation_loss}) starts to interfere with the main task training ($l_{normal}(\cdot)$) and fails to converge at the end of upstream training (Table~\ref{tab:accuracy_hyper-parameter}). For smile detection, $l_t(\cdot)$ still converges well (may be because the upstream model has enough capacity) and hence the inference effectiveness continues to increase as the increase of $\lambda$. 
%AUC scores increase as $\lambda$ increases when $\lambda \leq 5$, but decrease afterwards;. For age prediction, we also see a similar trend and the turning point is $\lambda = 10$. For smile detection, AUC scores increase as $\lambda$ increases for the considered range of $\lambda$. This is because the difference between the downstream models trained with and without the target property will be larger as $\lambda$ increases, making it easier for inference methods to capture a larger difference. But if $\lambda$ is too big and pursues a training goal beyond the model's capacity, the upstream training will not converge, and inference AUC scores will not increase or even decrease.  Figure~\ref{fig:hyper-parameter_lambda_black_box} in the appendix shows the results of the black-box meta-classifiers and have a similar trend as Figure~\ref{fig:hyper-parameter_lambda}. For gender recognition, we can still find $\lambda = 10$ as the turning point.

In Figure~\ref{fig:hyper-parameter_lambda}, although the choice of $\lambda$ does have some impact on the inference effectiveness, we find that our attack still works quite well for a wide range of $\lambda$ values. For example, 
%In Figure~\ref{fig:hyper-parameter_lambda} (the white-box inferences), 
for gender recognition, AUC scores are quite high and exceed 0.9 if $\geq 10$ samples are with the target property when the value of $\lambda$ is between 1 and 10; for the other two tasks, when the value of $\lambda$ is between 5 and 20, AUC scores also exceed 0.9 if $\geq 20$ samples are with the target property. We have similar observations as above (i.e., the trend of inference effectiveness as $\lambda$ changes and good attack performance for a wide range of $\lambda$) when we replace the white-box inference methods with black-box ones and details can be found in Figure~\ref{fig:hyper-parameter_lambda_black_box}.
%As to the black-box meta-classifiers in Figure~\ref{fig:hyper-parameter_lambda}, when $\geq 50$ samples have the target property, AUC scores exceed 0.9 if the value of $\lambda$ is between 1 and 10 for gender recognition and is between 5 and 20 for smile detection and age prediction. Those results suggest that it is easy for the attacker to find a suitable value for $\lambda$ to launch successful attacks.

\shortsection{Impact of \texorpdfstring{$\boldsymbol m$}{m}} \label{sec:study_m}
The hyperparameter $\boldsymbol m$ controls the location and number of activations selected for manipulation in Equation~\ref{eq:x_w_wo_activation}. We empirically find that, with the same size of activations $\|\boldsymbol m\|_{1}$, the location of $\boldsymbol m$ does not have a significant impact on attack effectiveness, and therefore, we fix the selection of manipulated activations to be the first $n_t$ activations (i.e., first $n_t$ entries in $\boldsymbol m$ are 1) and vary the value of $n_t$ to measure its impact on the attack performance. The rest of the experimental settings are the same as in Section~\ref{sec:eval_zero_activation_attack}. We choose the first $8, 16, 32$ and $64$ of the total 1$\,$280 activations as the secreting activations for the gender recognition task. For the smile detection and the age prediction tasks, we select the first $1, 4, 8,$ and 16 channels out of 256 channels as the secreting activations.
%Similarly to Section~\ref{sec:study_lambda}, we start with the settings the same as those in Section~\ref{sec:eval_zero_activation_attack} and vary the hyper-parameter $\boldsymbol m$. Since there are so many possibilities on the values of  $\boldsymbol m$, we can only consider the values from reasonable ranges. Specifically, for the gender recognition task, we select 8, 16, 32, and 64 activations out of 512 activations as the secreting activations separately; for the smile detection and the age prediction, we select 1, 4, 8, and 16 channels out of 256 channels as the secreting activations separately.

The inference methods adopted are the same as those in the study of the impact of $\lambda$  %Section~\ref{sec:study_lambda} 
and the white-box results are reported in Figure~\ref{fig:hyper-parameter_m}. From the figure, we observe that, in general, the inference effectiveness increases as we increase the number of selected activations (i.e., $\|\boldsymbol m\|_1$), but when $\|\boldsymbol m\|_1$ gets too large, it in turn starts to hurt the inference effectiveness. The possible reason is still similar to the one in the study of the impact of $\lambda$: 
%Section~\ref{sec:study_lambda}: 
initially, when more activations are selected for manipulation, the difference between the downstream models trained with and without the target property will be more significant, and makes the subsequent inference attacks more effective. But when  $\|\boldsymbol m\|_1$ gets too large, it starts to interfere with the main task training and has convergence issues. 
%the attack goal will be too hard to achieve in the upstream training. 
%We observe a similar trend as that in Section~\ref{sec:study_lambda}: in general, the inference performance increases with the number of activations selected by $\boldsymbol m$, but decreases if $\boldsymbol m$ chooses too many activations.
%The most likely explanation is: if more activations are manipulated, the difference between the downstream models trained with and without the target property will be bigger and will be easier for the inference methods to capture. But if the attacker tries to manipulate too much activations, the attack goal will be too hard to achieve in the upstream training.
From Figure~\ref{fig:hyper-parameter_m}, we also observe that the inference AUC scores remain high across all selections of $\boldsymbol m$. For example, AUC scores are all $>0.9$ when $\geq 20$ downstream training samples have the target property for gender recognition and smile detection and when $\geq 50$ downstream training samples are with the target property for age prediction. Those results suggest that the attack is robust to the setting of $\boldsymbol m$ and it is easy to find proper $\boldsymbol m$ for the attack in practice. Similar observations are also found when we replace the white-box inference methods with black-box ones (details in Figure~\ref{fig:hyper-parameter_m_black_box}). 
%for both the white-box inferences (Figure~\ref{fig:hyper-parameter_m}) and black-box meta-classifiers (Figure~\ref{fig:hyper-parameter_m_black_box}), AUC scores are all $>0.9$ when $\geq 20$ downstream training samples have the target property for gender recognition and smile detection and when $\geq 50$ downstream training samples are with the target property for age prediction. Those results suggest that the attack is robust to setting of $\boldsymbol m$ and the attacker can easily find a proper value for $\boldsymbol m$.

%\dnote{I don't think these hyperparameter explorations are very interesting, they feel mostly like filler material that should be in an appendix for completeness. Is there something interesting a reader should get from this subsection? For targeting a security conference, would be much more interesting to vary the threat model and see how sensitive results are to that.} \ynote{TODO FUTURE}

%\dnote{do we have any results that show how well the training works to train the target parameters? that would be the interesting ablation to do, to see if there is more room for improvement in improving the pretraining or in improving the inference tests, and to understand how sensitive we can make the pretraining+inference. I would also like to see how to use the results to predict the number of samples with the property. The results in the figures suggest we could do that, but no results on it, and hard to guess what the accuracy would be} \ynote{TODO FUTURE}

\begin{figure*}[ht]
    \centering
  \includegraphics[width=.95\linewidth]{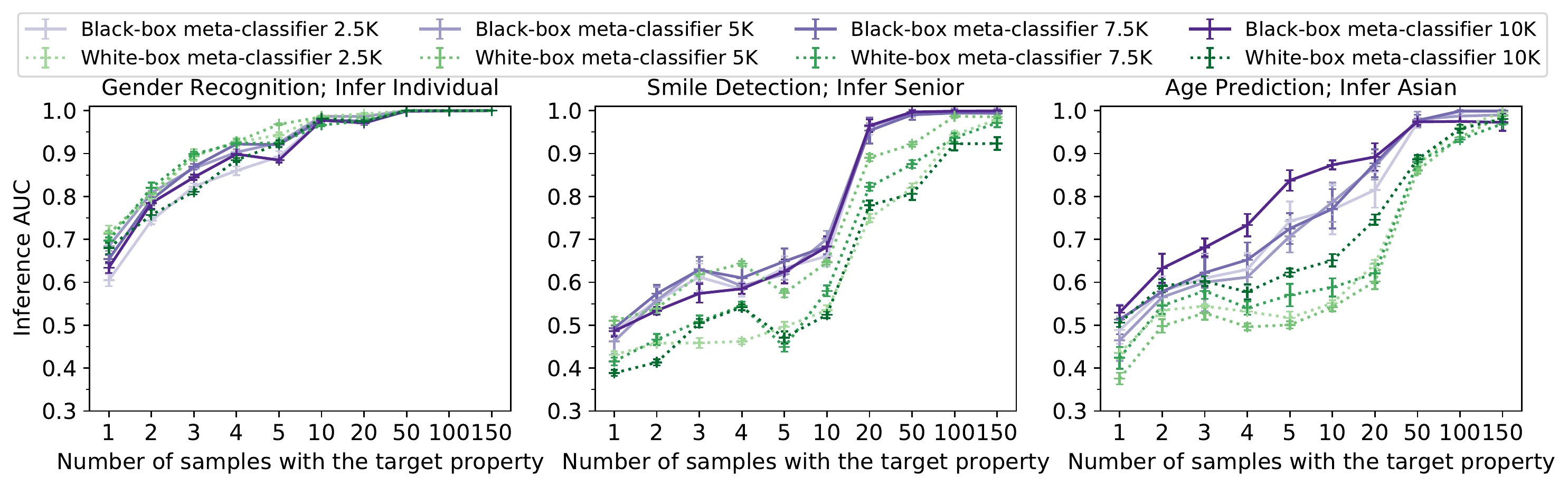} 
\caption{Inference AUC scores of meta-classifiers when the shadow models of the meta-classifiers are trained on datasets of different sizes. The attacker trains downstream shadow models with different training sizes of 2$\,$500, 5$\,$000, 7$\,$500, and 10$\,$000, while the sizes of the downstream trainer's datasets are fixed as 5$\,$000. % The inference targets for the smile detection and age prediction are senior people and Asian people respectively.
} 
\label{fig:unknown_size_results}
\end{figure*}

\begin{figure*}[tbp]
    \centering
    \includegraphics[width=.95\linewidth]{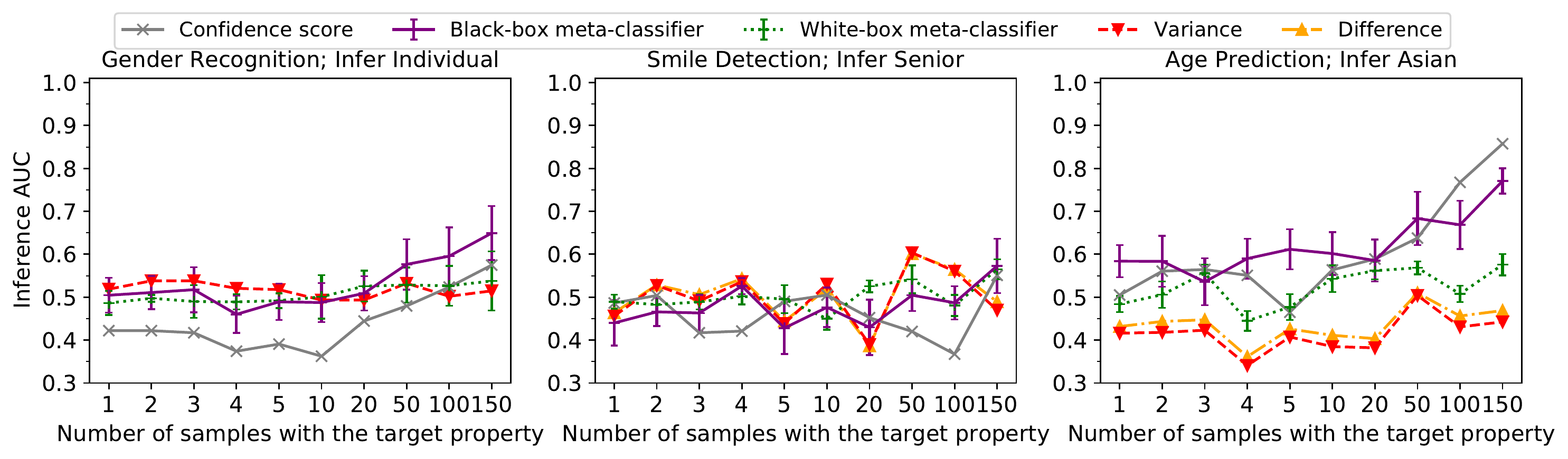}  
\caption{Inference AUC scores when upstream models are not trained with distribution augmentation (Appendix~\ref{sec:train_data_limitation}). % (Section~\ref{sec:attack_design}). 
All the downstream training sets have 5$\,$000 samples in these results. % The inference targets for the smile detection and age prediction are senior people and Asian people respectively. 
%and in the following figures in the main body of this paper, we will always use the same inference targets for those two tasks.
} 
\label{fig:distribution_augmentation}
% \vspace{-0.5em}
\end{figure*}

\subsection{Impact of the knowledge of the size of the downstream set }\label{sec:unknown_training_size}
%\dnote{shadow? I don't think we used this before} \ynote{we did used this before for the meta-classifiers} \ynote{found a better name for this subsetion}

%In Appendix~\ref{sec:eval_baseline} and 
In Section~\ref{sec:eval_zero_activation_attack}, when conducting property inference with meta-classifiers, the attacker trains shadow models using the same downstream training set size $n$ as the victim. In this section, we show that, for meta-classifier-based attacks, the knowledge of downstream training size used by the victim does not impact inference effectiveness much.

In the experiments, we fix the size of the victim training set to 5$\,$000 (i.e., $n=5\,000$) and vary the sizes of the (simulated) downstream training sets of the attacker. Specifically, we set the attacker training size to 
%we target the settings where the sizes of the downstream trainer's datasets are 5$\,$000 and vary the sizes of the downstream sets on which the attacker's shadow models will be trained. We set the attacker's downstream training sizes as 
2$\,$500, 5$\,$000, 7$\,$500, and 10$\,$000 separately and remaining experimental setups are kept the same as in Section~\ref{sec:eval_zero_activation_attack}.

%\dnote{don't understand why we would continue to consider white-box meta-classifier attacks - didn't we just show they are strictly worse than the other white-box attacks under the same threat model assumptions? only seems relevant to show that the black-box attacks are still effective without needing the assumption that the adversary knows the tuning set size. I don't think this is such an important result to make a subsection about it - maybe should be an appendix, or a shortsection in section that covers all the changes we consider in the threat model (but maybe this is the only one?) Since we don't see to have any discussion of black-box attacks in the previous section, could move this there.} \ynote{I think we can will revise this part according to this comment and add more different threat models in the future. Here, we still include the white-box meta-classifer to show all meta-classifiers are not sensitive to the training set. Our results imply that If there are more powerful meta-classifiers in the future, it will probably have the same property.} \ynote{TODO FUTURE}

Figure~\ref{fig:unknown_size_results} shows the inference results of the meta-classifier-based approaches. For both the white-box and black-box methods, varying the training set size has negligible impact on the inference performance: for the black-box approach, the purple lines stay very close to each other and the AUC scores all exceed 0.8 when $\geq 20$ samples out of the total 5$\,$000 samples have the target property and exceed 0.95 when $\geq 50$ samples are with the property. Similarly, for the white-box meta-classifiers approach, the green lines also stay close to each other and the AUC scores all exceed 0.9 when $\geq 100$ samples have the target property.
%Those results suggest that the training sizes of the shadow models will not affect the performance of the meta-classifiers too much.

%For the white-box meta-classifiers which are served as the baseline in the white-box scenarios, the results (green lines) show a similar trend as the purple lines but have exceptions when 150 samples have the target property for the smile detection task and age prediction task. The possible explanation for the exceptions is that the current state-of-the-art white-box meta-classifiers do not work very well for convolutional models (The downstream models of the smile detection and age prediction tasks are convolutional, while that of the gender recognition task contains only fully-connected layers). And those exceptions will not affect the inference, because we have inference methods (the parameter difference test and the variance test) that achieve better results for white-box scenarios.

\subsection{Importance of Distribution Augmentation} \label{sec:study_distribution_aug}
%\dnote{need more structure this, not just a list of experiments. The previous one was about changing the threat model - is that the only one about changing the threat model?}

% \begin{figure*}[tbp]
%     \centering
%     \includegraphics[width=.85\linewidth]{fig/attack/bn_zero_activation_no_distribution_augmentation_inference_5000.pdf}  
% \caption{Inference AUC scores when upstream models are not trained with distribution augmentation (Section~\ref{sec:attack_design}). All the downstream training sets have 5\,000 samples in these results. The inference targets for the smile detection and age prediction are senior people and Asian people respectively. 
% %and in the following figures in the main body of this paper, we will always use the same inference targets for those two tasks.
% } 
% \label{fig:distribution_augmentation}
% \end{figure*}

\begin{figure*}[ht]
    \centering
    \includegraphics[width=.95\linewidth]{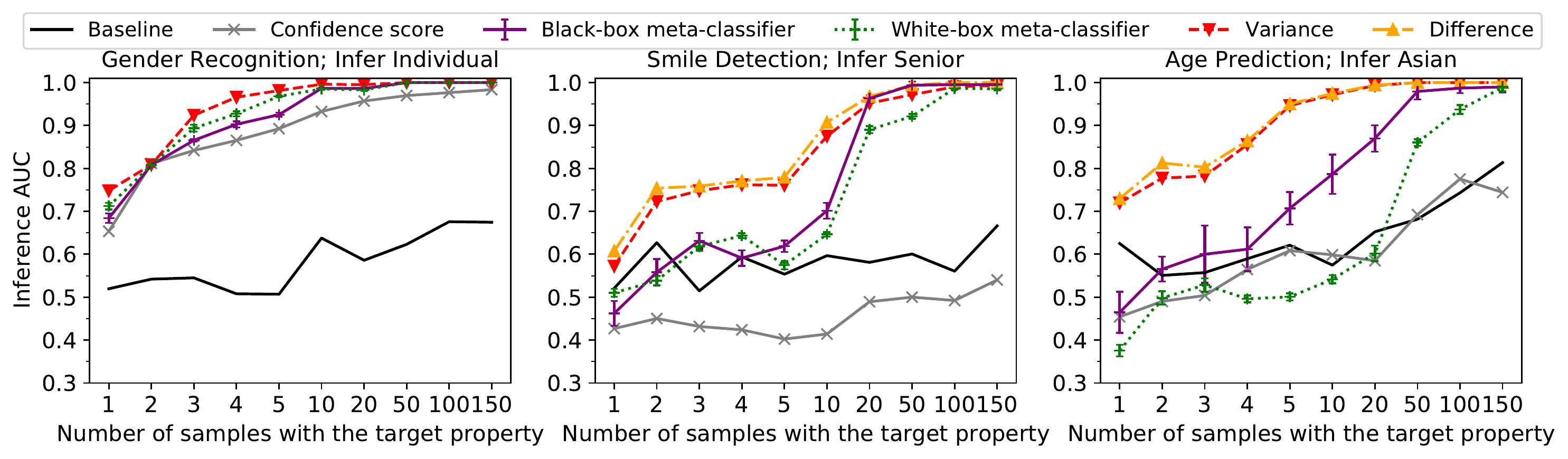}
\caption{Inference AUC scores when the upstream model is trained  %\dnote{not about the attack goals here, how is it trained?} 
with the attack method described in Section~\ref{sec:attack_design}.
Baseline scores (the \emph{baseline} lines) are the maximum of the AUC scores (of the three inference methods) of the baseline experiments in  Appendix~\ref{sec:eval_baseline}.
%In the gender recognition task, the downstream trainer reinitializes the classification module and the attacker does not know the initialization. Therefore, inference method based on parameter difference test is not applicable.
%The inference targets for the smile detection and age prediction are senior people and Asian people respectively.  
The inference of specific individuals for smile detection and age prediction are similarly successfully and found in Figure~\ref{fig:zero_activation_attack_results_t_individual} in the appendix. The downstream training sets have 5$\,$000 samples in the results,  and the results for the 10$\,$000 samples are in Figure~\ref{fig:zero_activation_attack_results}.
}
\label{fig:zero_activation_attack_results_5000}
\end{figure*}

% In the attack design (Section~\ref{sec:attack_design}),
In Appendix~\ref{sec:train_data_limitation},
%\dnote{ 
% should be more specific pointer - I find the paragraph in 4.1 about this, but
%it is quite vague and hard to understand what is actually done - if this is important enough to have experimental results on it, need to explain it more clearly and how you are doing it}) 
we introduce distribution augmentation for upstream training, which injects representative samples without the target property into the upstream training set to better achieve the attack goal described in Equation~\ref{eq:x_w_wo_activation}. Figure~\ref{fig:distribution_augmentation} shows the attack performance when we do not use distribution augmentation. The victim training set size is set to 5$\,$000 and other experimental setups are the same as those in Section~\ref{sec:eval_zero_activation_attack}. From the figure, we observe that AUC scores of attacks without distribution augmentation are all less than 0.86, and get even lower ($<0.7$) for gender recognition and smile detection. These scores are significantly lower than the results with distribution augmentation (details in Figure~\ref{fig:zero_activation_attack_results_5000} and \ref{fig:zero_activation_attack_results}). For example, with the augmentation, AUC scores all exceed 0.9 if more than 20 samples are with the target property and the importance of distribution augmentation is thus apparent.

\subsection{AUC values < 0.5} \label{sec:auc<0.5}
We observe that a few attack settings have AUC scores consistently below 0.5. Those rare abnormal AUC scores mainly occur for black-box methods against normal pretrained models (e.g.,
the confidence score test and black-box meta-classifier for the gender recognition with 10$\,$000 downstream samples in Figure~\ref{fig:baseline_results}.) For the confidence score test, by manual inspection, we find its working assumption is not satisfied by the downstream models fine-tuned from normal pretrained models in some settings. The confidence score test assumes models trained with the property perform better on samples with the property than those trained without the property, but an opposite pattern is observed for the queried downstream models. 
As for black-box meta-classifiers, we observe the anomalies happen when the inference tasks are too challenging and the meta-classifiers cannot obtain meaningful information but overfit to the training set (despite early stopping). Specifically, AUC scores are high ($>0.75$) on the training set, $\sim0.5$ on the validation set, and show anomalies ($<$ 0.5) on the test set. We note that the gap between the validation set and the test set is large because they are trained differently. When training downstream models with the target property for the training and validation set, we randomly sample 1-170 samples with the property each time to simulate the real-world case (discussed in Appendix~\ref{sec:details-of-downstream-training}), while for the test set, we randomly sample fixed number of samples with the property for each AUC computation (e.g., 1, 2, ..., 150) to show the trend. %In most of these rare cases, AUC scores still remain low even if the attacker flips predictions ($1 - AUC< 0.7$).  
We reemphasize that those anomalies mainly happen in the non-manipulation settings because of the limitation of inference methods on normal pretrained models when the inference tasks are too challenging. Our proposed manipulation (e.g., providing stronger signal) lowers the difficulty of those challenging cases and leads to better/normal results.

\subsection{Inferring Multiple Properties Simultaneously} \label{sec:multiple_properties}
\begin{figure}[ht]
    \centering
  \includegraphics[width=.45\textwidth]{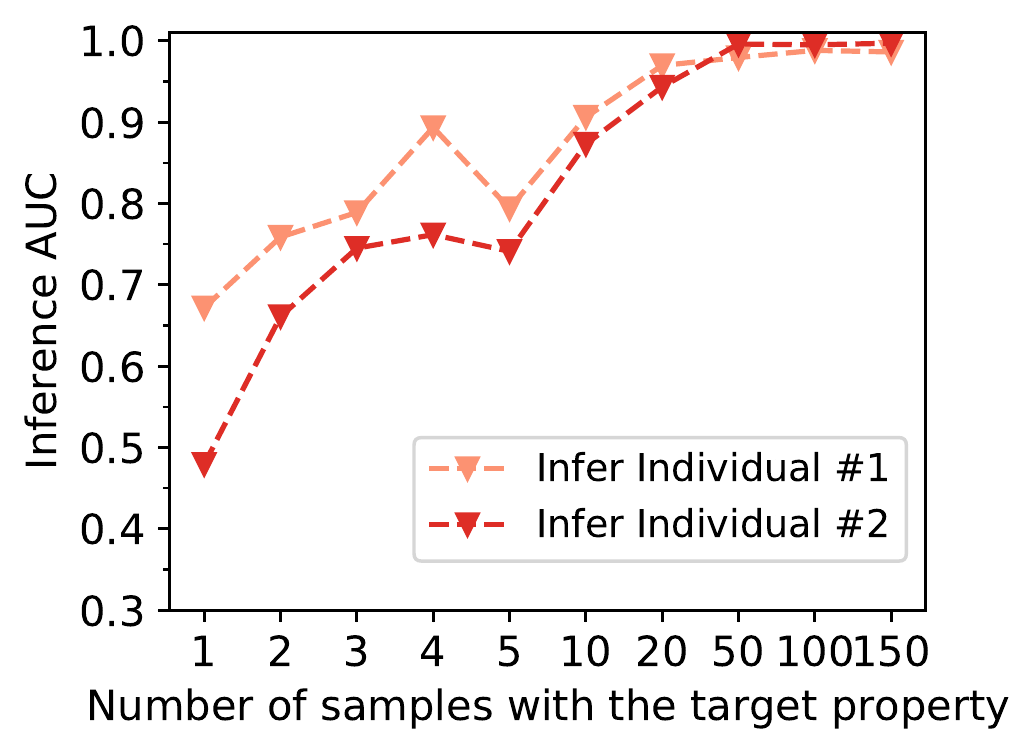} 
\caption{Inference AUC scores when considering multiple properties simultaneously. The inference task is to infer two individuals in the gender recognition setting. The downstream set has 5$\,$000 samples.
} 
\label{fig:infer_multiple_properties}
\end{figure}

In this section, we demonstrate that the attack described in Section~\ref{sec:attack_design} can be extended to infer multiple target properties simultaneously. The method is to simply associate different secreting parameters with each property. We conducted experiments using the gender recognition setting with some modifications. The new target properties are the two individuals with the most samples in VGGFace2. In the upstream training, we inject 285 and 257 samples with the property into the upstream training set for the two individuals respectively; we also inject 1$\,$425 samples without the target properties (distribution augmentation in Appendix~\ref{sec:train_data_limitation}). For each property, the number of scereting activations is 8 (i.e., $\|\boldsymbol m\|_1=8$). For the downstream training, the candidate set has 250 samples for each target property and 200$\,$000 samples without the target properties.
The rest settings are the same as those in Appendix~\ref{sec:hyper-parameter-setup-zero-activation}. 
The manipulation does not affect the accuracy of the main tasks too much (accuracy drop less than 0.6\%). 
The inferences are also highly successful. Figure~\ref{fig:infer_multiple_properties} summarizes the results of the variance test in discriminating downstream models trained with a target property from those trained without target properties. The results show that AUC scores exceed 0.85 when $\geq 10$ out of 5$\,$000 samples are with the property, and are higher than 0.95 when $\geq 50$ samples have the property.

\subsection{Details on Anomaly Detection for Zero-Activation Attack}
\label{sec:defense_details}
We consider three common anomaly detection methods: K-means~\cite{jain1999data}, PCA~\cite{abdi2010principal} and Spectre~\cite{hayase2021spectre}, where Spectre is the current state-of-the-art. 
K-means leverages the k-means clustering technique to identify outliers while PCA leverages principal component analysis to identify the outliers. Spectre is an improved version of PCA and works much better than PCA when the attack signature is weak (i.e., the distributional difference is small)~\cite{hayase2021spectre}.
%originally designed to detect the backdoored samples (can still be viewed as outliers) in model training to defend against backdoor attacks. 
%It works by first finding the target class of the backdoor attack, and then performing anomaly detection with the activations of the samples labeled into the target class. In our case, since samples with the target property can span different output classes, we skip the step of finding the target class and directly conduct anomaly detection with the feature extractor's outputs on the downstream set. 
When conducting the anomaly detection, following the common setup in Hayase et al.,~\cite{hayase2021spectre}, we filter out $1.5 n_t$ ($n_t$ is number of samples with target property) samples, simulating the scenario where the defender does not know the exact $n_t$, but is able to roughly estimate its value and attempt to find most of them.
%\dnote{need to explain how these relate to our problem - the do not seem like appropriate defenses for our setting, so if it makes sense to use them need to explain why}

% \shortsection{Defense Performance Against Zero-Activation Attack}

\begin{figure*}[htbp]
    \centering
    \includegraphics[width=.95\linewidth]{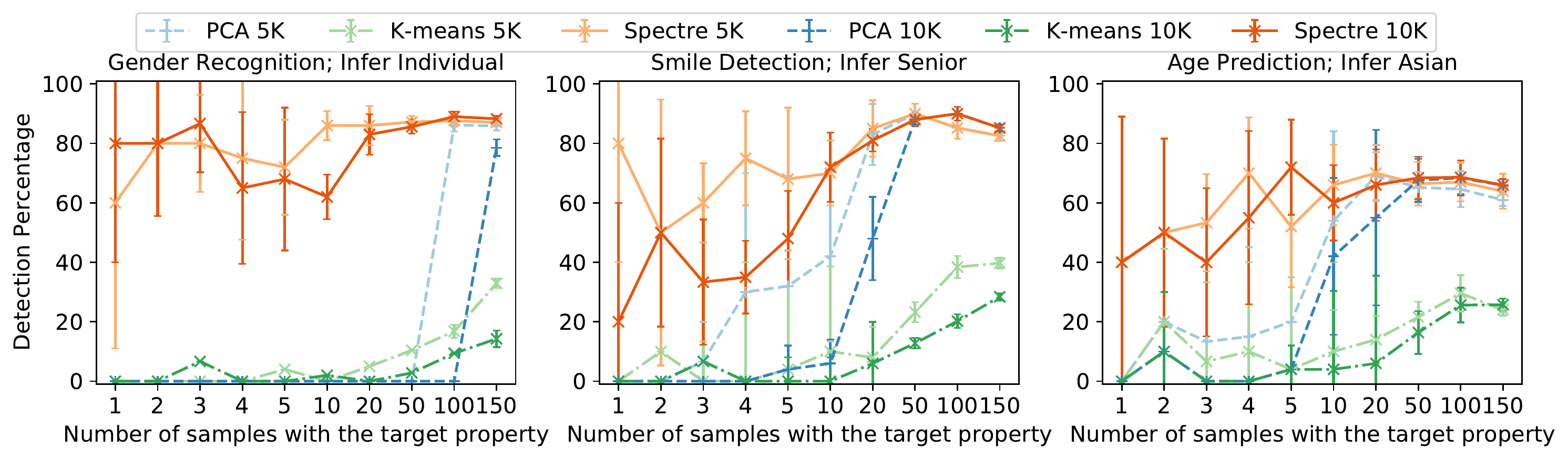} 
\caption{Percentage of samples with the target property detected by the anomaly detection for the zero-activation attack.
Similar to ~\cite{hayase2021spectre}, we filter out $n \times 1.5$ samples with anomaly detection, where $n$ is the number of samples in downstream training data with the target property. We report the number of samples with the target property filtered out divided by $n$ as the \emph{Detection Percentage}; values are averaged (with standard deviation) over 5 runs of anomaly detection.
The `5K' lines report detection results on the settings with 5$\,$000 total samples, while the `10K' lines report for 10$\,$000 total samples.
%Total number of filtered samples are $1.5$ times the number of actual target samples and the detection percentage is the fraction of actual target samples that are filtered out. We repeat the experiments for 5 times and report the average value and standard deviation.
%The `5K' and `10K' lines report for settings of downstream training sets of 5000 and 10 000 total samples, respectively.
\label{fig:anomaly_detection_zero_activation_attack} }
\end{figure*}

\shortsection{Results of Anomaly Detection} 
\label{sec:defense-zero-activation}
We show the detection performance in Figure~\ref{fig:anomaly_detection_zero_activation_attack}. 
The results show that conducting anomaly detection can filter out majority of samples with the target property in the downstream set and hence, increase the chance of detecting the manipulation.
% lower the effectiveness of the inference \todo{change the story}.
For example, the Spectre defense can filter out 80\% of the samples with the target property in most cases for gender recognition and smile detection, and 60\% for age prediction. Anomaly detection effectively finds samples with the target property
% detects zero activation attacks
because the attack mainly focuses on improving attack effectiveness by increasing the distinction between samples with and without property, which makes the attack signature of samples with property much stronger.  After finding the possible samples with the target property, the defender can then inspect those samples, and try to find the commonalities and then identify the potential target property. Since the process of finding commonalities in the outliers reported by anomaly detection could be trivial (e.g., most samples have the same property or abnormal activations), we do not perform actual experiments for this part. In Section~\ref{sec:stealthier-design-methods}, we propose a stealthier design, in which anomaly detection cannot reliably detect samples with the target property and thus cannot find the manipulation.

\subsection{Experimental Setup of Stealthier Attacks}\label{sec:experimental-setup-stealthier-attacks}
In Section~\ref{sec:stealthier-design-results}, when preparing upstream models, for $\boldsymbol m$, we randomly select 16 activations out of total 1$\,$280 for the gender recognition and also select 196 activations out of total 50$\,$176 for smile detection and age prediction.
In practice, the total number of channels in convolutional kernels is not very large and therefore, the defender may still be able to brute-force the manipulated activations if $
\boldsymbol m$ is chosen only at the channel level. Thus, we also choose to select secreting activations directly for tasks where the first layer of the downstream model is convolutional, which may reduce some of the attack effectiveness.
%Another small change is, if the first layer of downstream model is convolutional, we still generate $\boldsymbol m$ for the secreting activations (which may lead to some slight loss in attack effectiveness) instead of generating it for the channels. We do this because we are dealing with an active defender who knows the attack method and generating $\boldsymbol m$ this way can significantly increase the difficulty of defender guessing the secreting activations (more details in the analysis of stealthiness below). 
For $\lambda$, we prefer a larger value for better inference effectiveness while still evading anomaly detection. Therefore, we performed a linear search starting from 1 and incrementing it by 0.5, and terminating when the attack can no longer evade the mentioned anomaly detection methods. With this strategy, we set $\lambda=2$ for gender recognition, $\lambda=1.5$ for smile detection and age detection when the inference targets are senior people and Asian people respectively, and $\lambda=1$ for smile detection and age detection when the inference targets are specific individuals. $\alpha$, $\beta$, and $\gamma$ are all set to be 1 in the experiments. 

\subsection{Adaptive Activation Distribution Checking} \label{sec:adaptive_activation_distribution_checking}
The activation distribution checking method needs to be adjusted based on the specific attack method used. Using the modified loss design in Section~\ref{sec:stealthier-design-methods}, our stealthier attack can automatically evade distribution checking of abnormal zeros, as the secreting activations of samples without target property are also non-zero. Hence, we need to design adaptive detection based on activation distribution checking for the modified attack loss. 

With the modified attack loss, we find that activations of samples with the property mixes well with ones without the property, and we fail to find a principled method to distinguish their distribution using the overall activations. Because of the design of the attack loss, the main distributional difference %always
comes from the distributional difference in the secreting activations for samples with and without property (i.e., distributional difference is most significant when we only measure secreting activations), to make progress, we assume the defender will follow a two-stage strategy of
first identifying the selected secreting activations and then identifying the distributional difference in the potential secreting activations, with a hope that the distributional difference is significant enough to be detected\footnote{We do not exclude the possibility of identifying the distributional difference by still checking the overall distribution, and leave further exploration of such detection strategies as future work.}. 

Since $\boldsymbol m$ is randomly generated with proper number of nonzeros, the brute-force strategy for identifying $\boldsymbol m$ is computationally infeasible. For example, for gender recognition experiments, defenders have to try a total of $\binom{1,280}{16}$ ($>2e36$) forms of $\boldsymbol m$ (i.e., $\|\boldsymbol m\|_1=16$ for a total of 1,280 activations). Therefore, alternatively, we present two methods that attempt to approximately identify $\boldsymbol m$ with the hope that the approximately well identified $\hat{\boldsymbol m}$ still preserves the significant distributional difference of $\boldsymbol m$. 
%In practice, the defender may not need to exactly identify $\boldsymbol{m}$, instead we assume approximately recovering most of the activations in $\boldsymbol{m}$ may also be sufficient~\footnote{Since the activation checking based defense is mostly a proof-of-concept, we assume identifying the secreting activations with sufficiently high detection rate implies the defender can also easily spot if the given model is manipulated or not. Similarly, if the detection rate is low, we assume the defender cannot confidently judge if the model is manipulated. We leave in-depth investigation of defenses along this path as the future work.}. Therefore, we also design approaches that attempt to approximately identify most of the actual secreting activations. 
The two methods we design are based on the fact that: 1) samples with the target property are rare for practically interesting settings, and 2) in the modified loss design, secreting activations of samples without the property are smaller in magnitude than the ones of samples with the property. Therefore, if we randomly feed inputs to the model, most of the inputs are without property and hence, their corresponding secreting activations should be smaller. With these two principles, we design two detection methods: the first one averages the outputs of each activation for all the fed inputs and treats activations with smaller average values as the potential secreting activations (\emph{average value based detection});
the second approach handles individual input separately and identifies potential secreting activations for each of them, and then returns the intersection for all the potential secreting activations identified (\emph{intersection based detection}). Empirically, we find that both approaches cannot identify the secreting activations well (details are shown below) %(\autoref{sec:stealthier-design-results}) 
and hence did not further explore how to check distributional difference on the identified secreting activations in this paper. 

\shortsection{Experimental Settings}
To evaluate the performance of \textit{average value based detection}, we measure the detection rate, which is the fraction of actual secreting activations in identified potential activations.  For the \textit{intersection based method}, since the size of final returned secreting activations can vary (due to intersection over multiple inputs) for different settings, we evaluate the defense performance by reporting their F1-score (viewing actual target as the positive class and others as negative). When running these two detections, we consider an idealized scenario for the defender, where all the randomly sampled inputs are without target property and so, their secreting activations are even smaller for manipulated models and are easier to be detected by the defender.

Specifically, for average value based detection, we choose $n \times 1.5$ activations that have the smallest average values as the identified possible secreting activations ($n_{ip}$), where $n$ is the number of actual secreting activations ($n=\|\boldsymbol m\|_1$). We report the number of  identified actual secreting activation ($n_{ia}$) divided by $n$ as the detection rate.  For intersection based detection, the $n_{ip}$ of this method is the number of activations remained after intersection operations, and we cannot precisely control this number. Therefore, only reporting the detection rate like the average value based detection could introduce bias, and we use the F1-score as the metric instead, where the precision is defined as $\frac{n_{ia}} {n_{ip}}$ and the recall is defined as $\frac{n_{ia}} {n}$. And for this detection method, for each sample, we also need to select some activations that have the smallest values as the inputs for conducting the intersection operation. We tried many choices for the number of those activations, and find that choosing $n \times 5$ smallest activations for each sample achieves the best F1-score. In the experiments, we tried to use 100, 500, 1$\,$000, 2$\,$000, 4$\,$000, 8$\,$000, 10$\,$000 samples to generate  activations values, separately. For each setting, we repeat  each detection 5 times and calculate the average value of the detection rate or F1-score. 

\shortsection{Detection Results} Empirically, we find that the two approaches cannot sufficiently identify the secreting activations --- the detection rate of secreting activations of the first method is less than 11.3\% for gender recognition and is less than 1.5\% for smile detection and age prediction for all settings; the F1-score of the secreting activation detection of the second method is less than 0.009 for all settings. In fact, using the second approach, the returned secreting activations are empty sets in most settings, implying the difficulty of identifying the secreting activations by simply checking the magnitude. Overall, the detection performances of both approaches are low and better detection methods are needed for identifying $\boldsymbol{m}$ in the future. %More details of the experiments and results are in Section~\ref{sec: details_distribution_checking} in the appendix.

\begin{figure*}[ht]
    \centering
    \includegraphics[width=.75\linewidth]{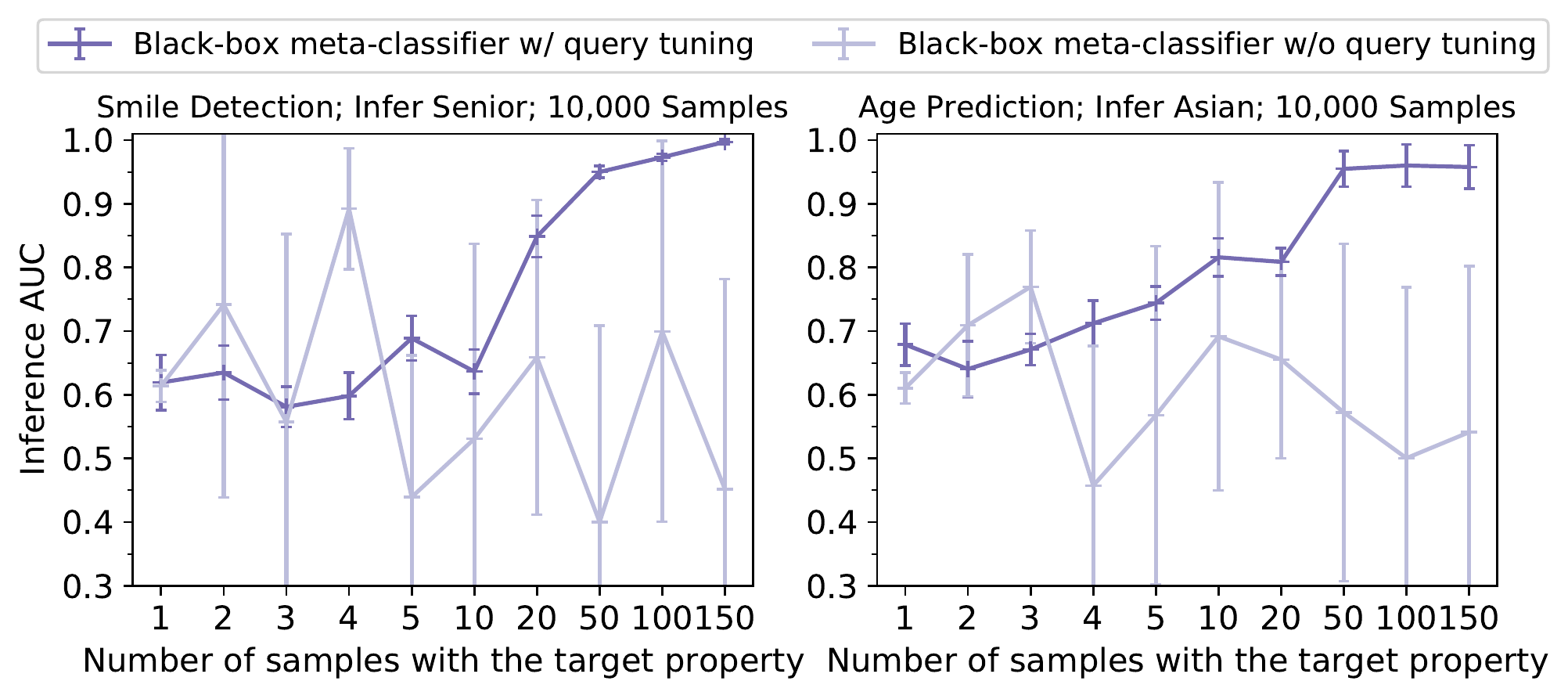}
\caption{Inference AUC scores of black-box meta-classifiers equipped with and without query tuning. 
We reuse the upstream and downstream models trained in Figure~\ref{fig:zero_activation_attack_results}.
}
\label{fig:query_tuning}
\end{figure*}

\begin{figure*}[tbp]
    \centering
    \includegraphics[width=.75\linewidth]{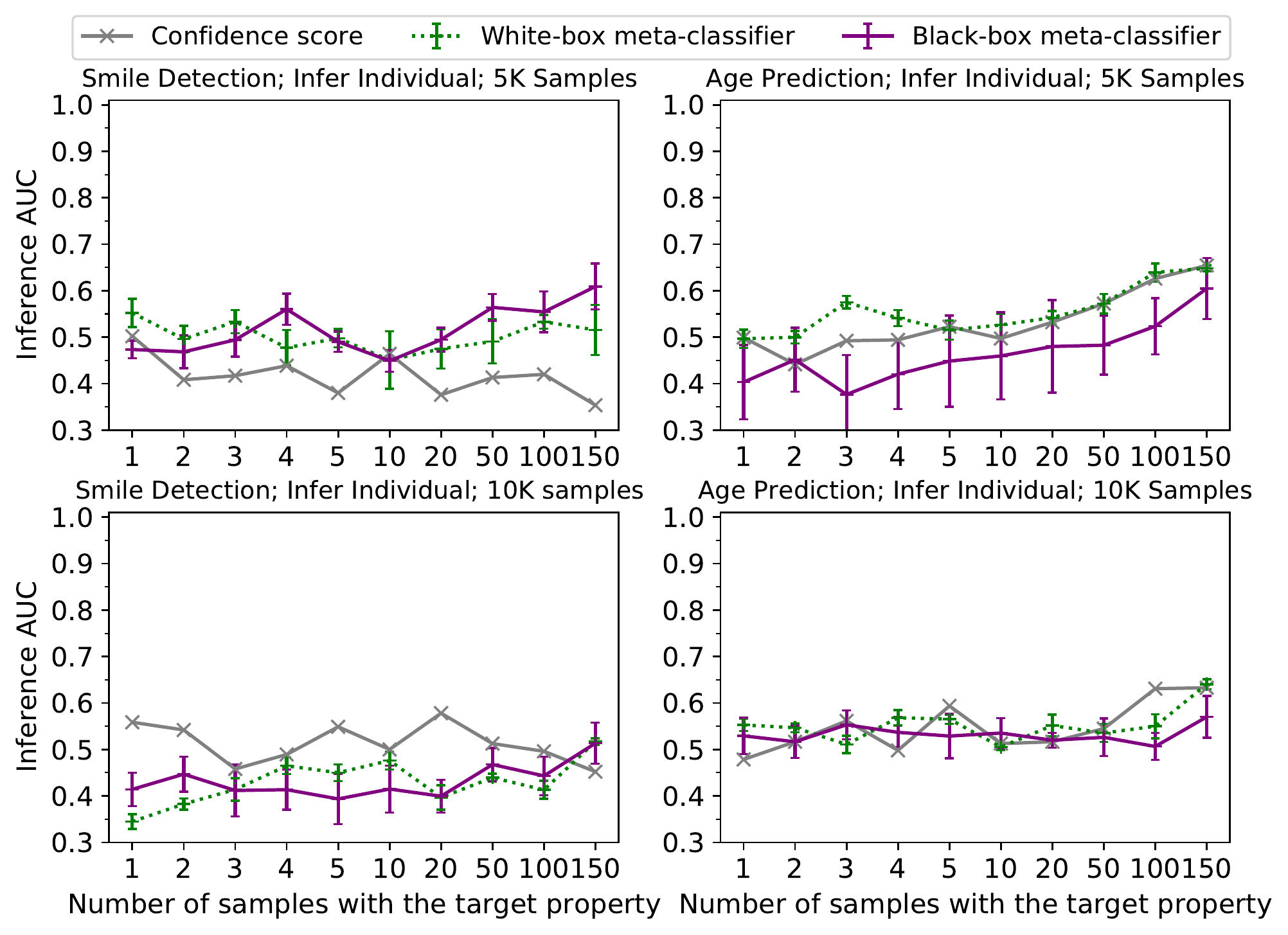}  
    \caption{Inference AUC scores when the upstream model is not trained with attack goals. The first and second rows show results when downstream training sets contain 5$\,$000 and 10$\,$000 samples respectively. The inference targets are specific individuals for smile detection and age prediction; the results of other inferences show a similar trend and are found in Figure~\ref{fig:baseline_results}.}
    \label{fig:baseline_results_t_individual}
\end{figure*}

\begin{figure*}[tbp]
    \centering
    \includegraphics[width=.75\linewidth]{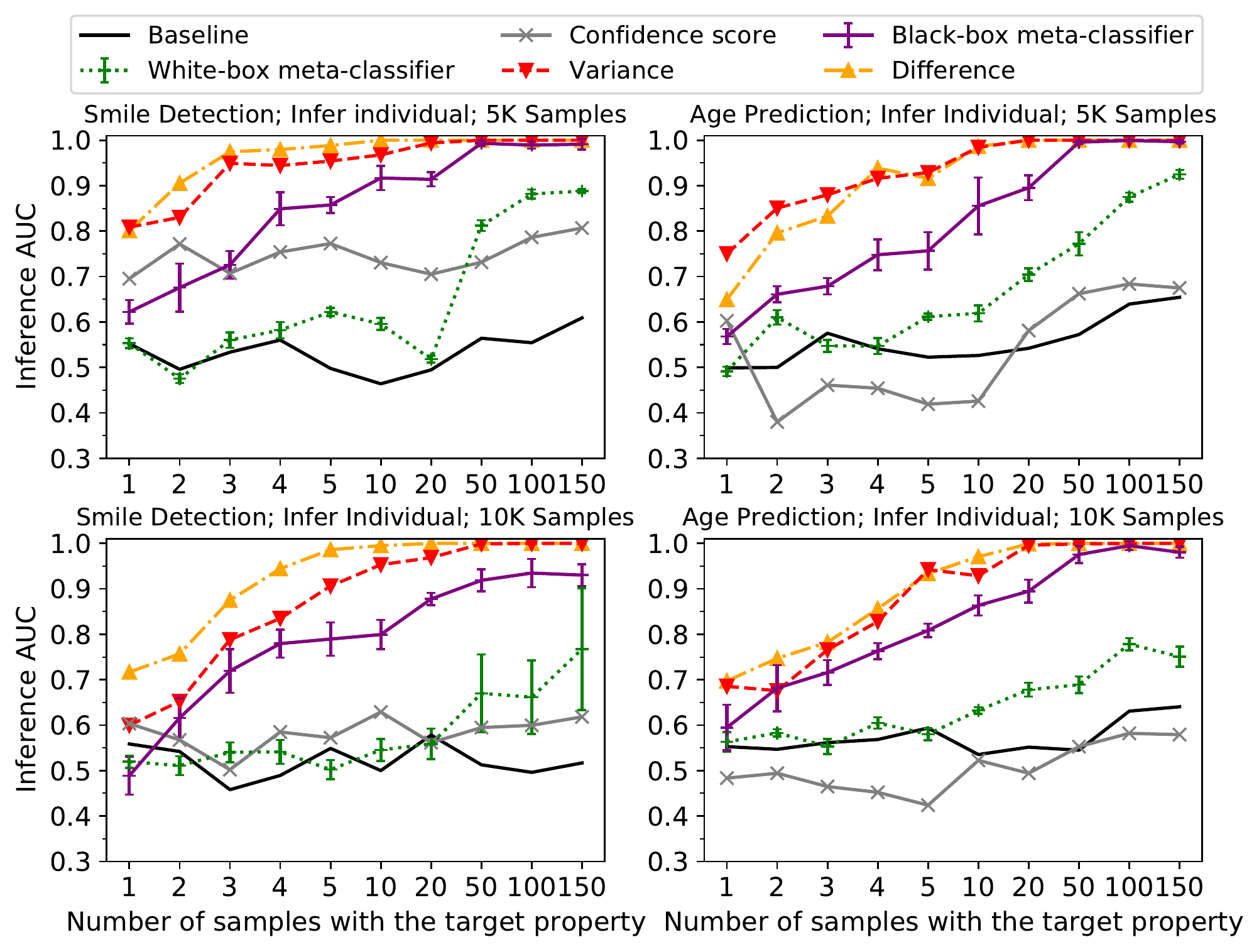}  
\caption{Inference AUC scores when the upstream model is trained with the attack goals described in Section~\ref{sec:attack_design}. The first and second rows show results when downstream training sets contain 5$\,$000 and 10$\,$000 samples respectively.
%The downstream training sets all have 5$\,$000 samples in the results.
The inference targets are specific individuals for smile detection and age prediction; the results of other inferences show a similar trend and are found in Figure~\ref{fig:zero_activation_attack_results}.
}
\label{fig:zero_activation_attack_results_t_individual}
\end{figure*}

\begin{figure*}[ht] %[htbp]
    \centering
    \includegraphics[width=.95\linewidth]{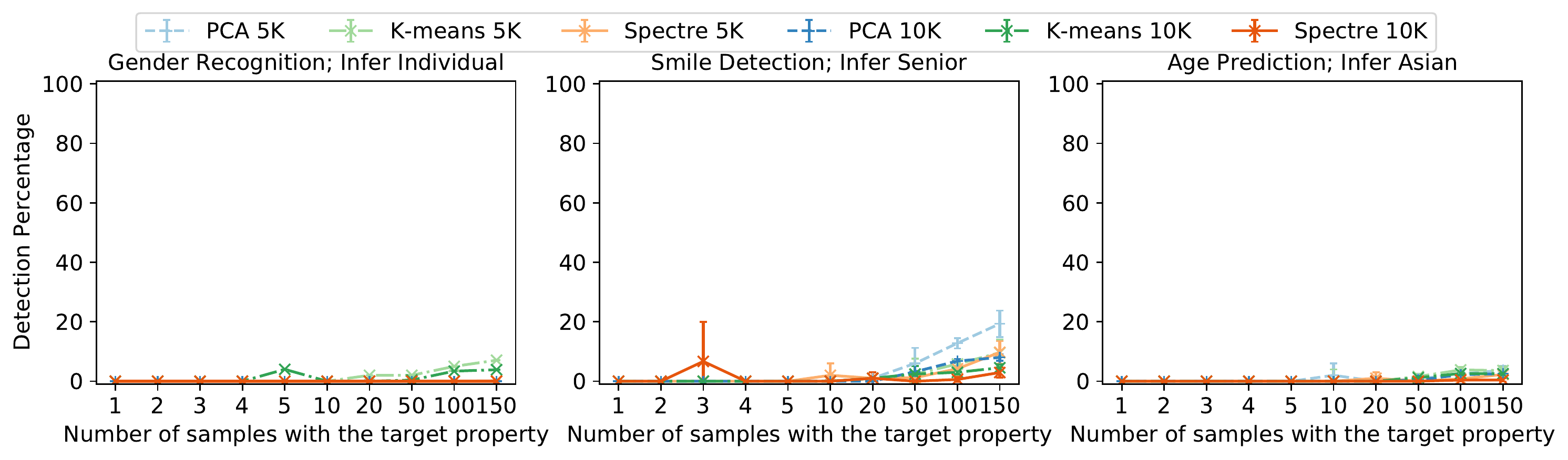} 
\caption{Percentage of samples with the target property detected by the anomaly detection for the stealthier attack. 
Similar to ~\cite{hayase2021spectre}, we filter out $n \times 1.5$ samples with anomaly detection, where $n$ is the number of samples in downstream training data with the target property. We report the number of samples with the target property filtered out divided by $n$ as the \emph{Detection Percentage}; values are averaged (with standard deviation) over 5 runs of anomaly detection.
The `5K' lines report detection results on the settings with 5$\,$000 total samples, while the `10K' lines report for 10$\,$000 total samples. Inference targets for smile detection and age prediction are senior people and Asian people respectively; results for the inference of specific individuals follow similar trends (Figure~\ref{fig:anomaly_detection_stealthier_attack_t_individual}).
\label{fig:anomaly_detection_stealthier_attack} }
%\vspace{-1em}
\end{figure*}

\begin{figure*}[htbp]
    \centering
    \includegraphics[width=.95\linewidth]{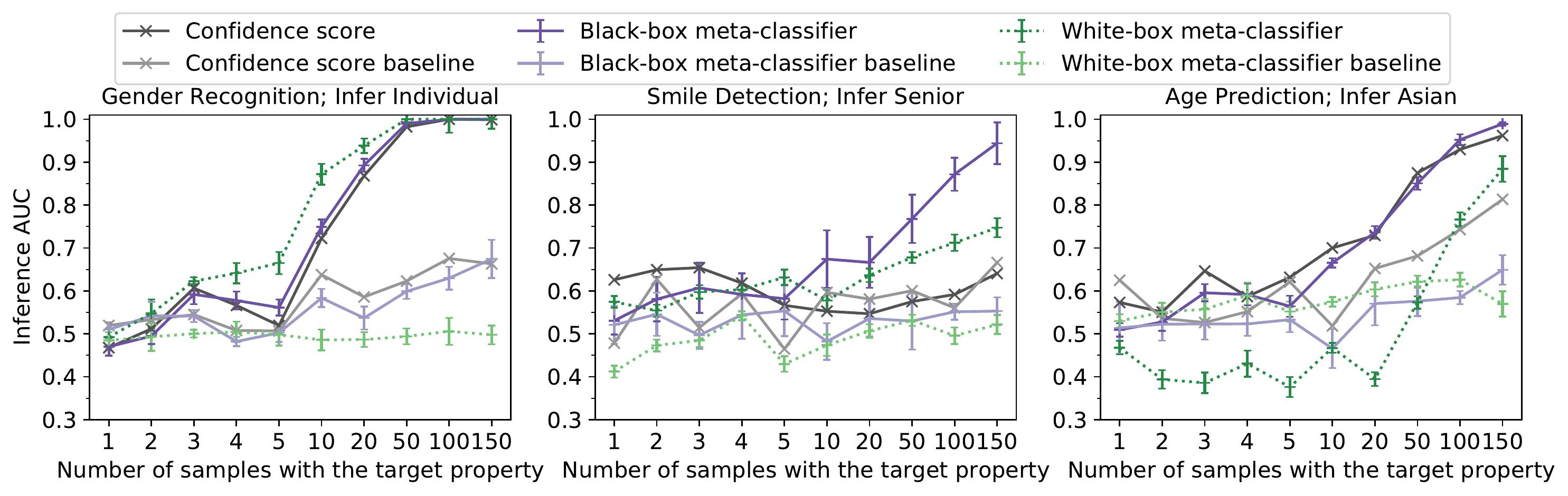} 
\caption{Inference AUC scores of the stealthier design. Since the secreting activations are no longer zero, the inference methods based on difference or variance tests are no longer applicable. 
Inference targets for the smile detection and age prediction are senior people and Asian people respectively; inference of specific individuals also shows improvement compared to the baseline settings (Figure~\ref{fig:stealthier_attack_results_t_individual}). The downstream training sets have 5$\,$000 samples in the results; results for 10$\,$000 samples show similar trends and are in Figure~\ref{fig:stealthier_attack_results_10000}.
\label{fig:stealthier_attack_results_5000}}
\end{figure*}

\begin{figure*}[htbp]
    \centering
    \includegraphics[width=.75\linewidth]{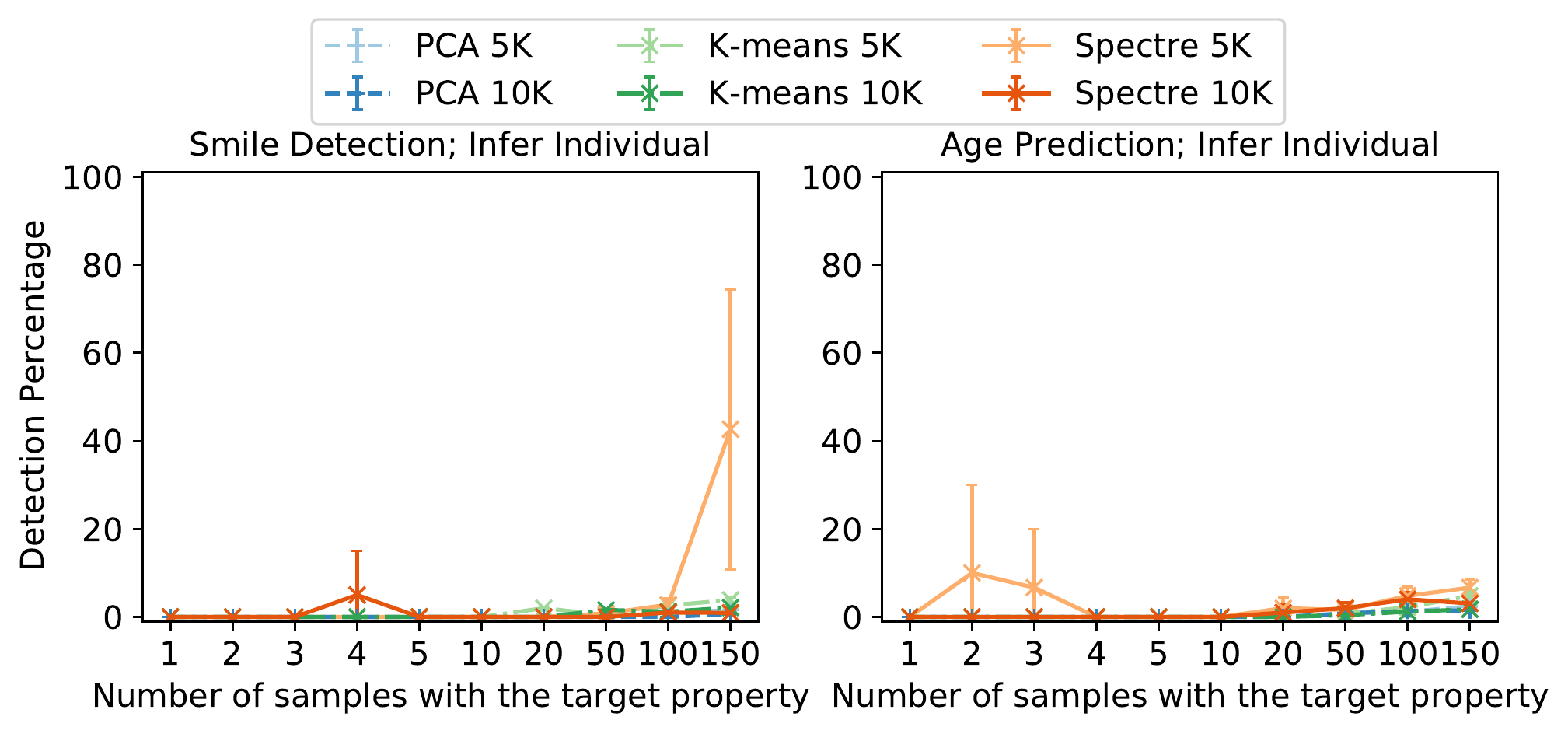} 
    \caption{Percentage of samples with the target property detected by anomaly detection for the stealthier attack. The inference targets are specific individuals for smile detection and age prediction; the results of other inferences show a similar trend and are found in Figure~\ref{fig:anomaly_detection_stealthier_attack}.
    \label{fig:anomaly_detection_stealthier_attack_t_individual} }
\end{figure*}

\begin{figure*}[htbp]
    \centering
    \includegraphics[width=.75\linewidth]{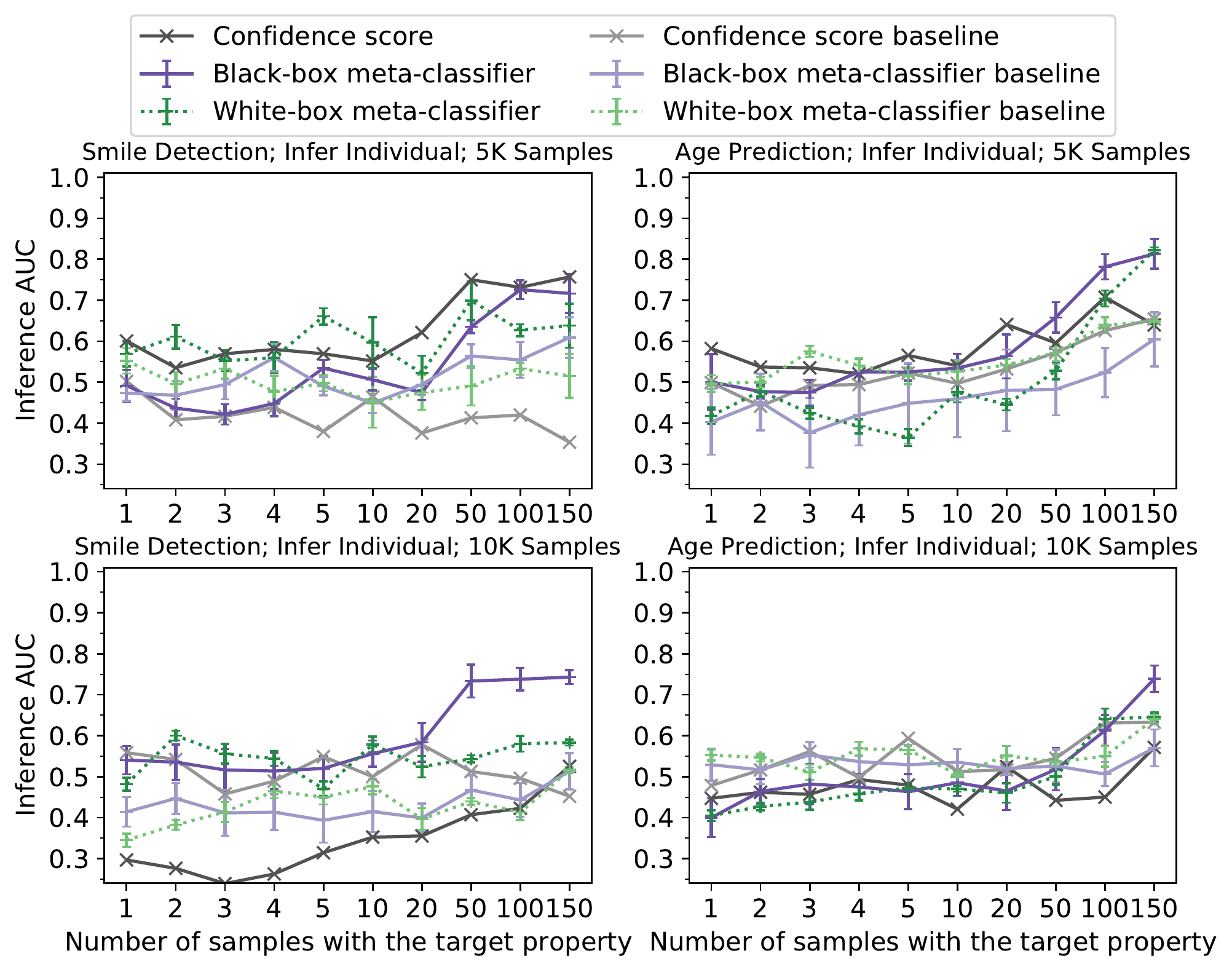} 
    \caption{Inference AUC scores of the stealthier attack. % The downstream training sets all have 5$\,$000 samples in the results. 
    The first and second rows show results when downstream training sets contain 5$\,$000 and 10$\,$000 samples respectively.
    The inference targets are specific individuals for smile detection and age prediction; the results of other inferences show a similar trend and are found in Figure~\ref{fig:stealthier_attack_results_10000}.
    \label{fig:stealthier_attack_results_t_individual}}
\end{figure*}

\end{document}